\documentclass[preprint,12pt]{elsarticle}   

\usepackage{amsmath,amssymb,amsthm,bm}

\usepackage{graphicx}
\usepackage{subcaption}
\usepackage{booktabs, multirow, longtable, array, adjustbox}
\usepackage{algorithm}
\usepackage{algpseudocode}
\usepackage{listings}
\usepackage{listingsutf8}
\usepackage{float}
\usepackage{lscape}

\usepackage[table,dvipsnames]{xcolor}
\definecolor{lightgreen}{RGB}{220,255,220}
\definecolor{lightred}{RGB}{255,220,220}
\definecolor{lightgray}{gray}{0.90}

\usepackage{hyperref}

\usepackage[
  draft,                 
  authormarkup=none
]{changes}

\newtheorem{definition}{Definition}

\journal{Computer Networks}

\begin{document}

\begin{frontmatter}



\title{Symbiotic Agents: A Novel Paradigm for Trustworthy AGI-driven Networks}


\author[inst1]{Ilias~Chatzistefanidis}
\ead{ilias.chatzistefanidis@eurecom.fr}

\author[inst1,inst2]{Navid~Nikaein}
\ead{navid.nikaein@eurecom.fr, navid.nikaein@bubbleran.com}

\affiliation[inst1]{organization={EURECOM, Sophia-Antipolis}, country={France}}
\affiliation[inst2]{organization={BubbleRAN, Sophia-Antipolis}, country={France}}

\begin{abstract}

Large Language Model (LLM)-based autonomous agents are expected to play a vital role in the evolution of 6G networks, by empowering real-time decision-making related to management and service provisioning to end-users.
This shift facilitates the transition from a specialized intelligence approach, where artificial intelligence (AI) algorithms handle isolated tasks, to artificial general intelligence (AGI)-driven networks, where agents possess broader reasoning capabilities and can manage diverse network functions.
In this paper, we introduce a novel agentic paradigm that combines LLMs with real-time optimization algorithms towards Trustworthy AI, defined as \textit{symbiotic agents}. 
Optimizers at the LLM's input-level provide bounded uncertainty steering for numerically precise tasks, whereas output-level optimizers supervised by the LLM enable adaptive real-time control.
We design and implement two novel agent types including: (i) Radio Access Network (RAN) optimizers, and (ii) multi-agent negotiators for Service-Level Agreements (SLAs). 
We further propose an end-to-end architecture for AGI-driven networks and evaluate it on a 5G testbed capturing channel fluctuations from moving vehicles.
Results show that symbiotic agents reduce decision errors fivefold compared to standalone LLM-based agents, while smaller language models (SLM) achieve similar accuracy with a 99.9\% reduction in Graphical Processing Unit (GPU) resource overhead and in near-real-time (near-RT) loops of $82~ms$. 
A multi-agent demonstration for collaborative RAN on the real-world testbed highlights significant flexibility in service-level agreement and resource allocation, reducing RAN over-utilization by approximately 44\%.
Drawing on our findings and open‑source implementations, we introduce the symbiotic paradigm as the foundation for next‑generation, AGI‑driven networks—systems designed to remain adaptable, efficient, and trustworthy even as LLMs advance. A live demo is presented here \href{https://www.youtube.com/watch?v=WQv61z1deXs&ab_channel=BubbleRAN}{$https://www.youtube.com/watch?v=WQv61z1deXs\&ab\_channel=BubbleRAN$}

\end{abstract}

\begin{graphicalabstract}
\includegraphics[width=0.99\textwidth]{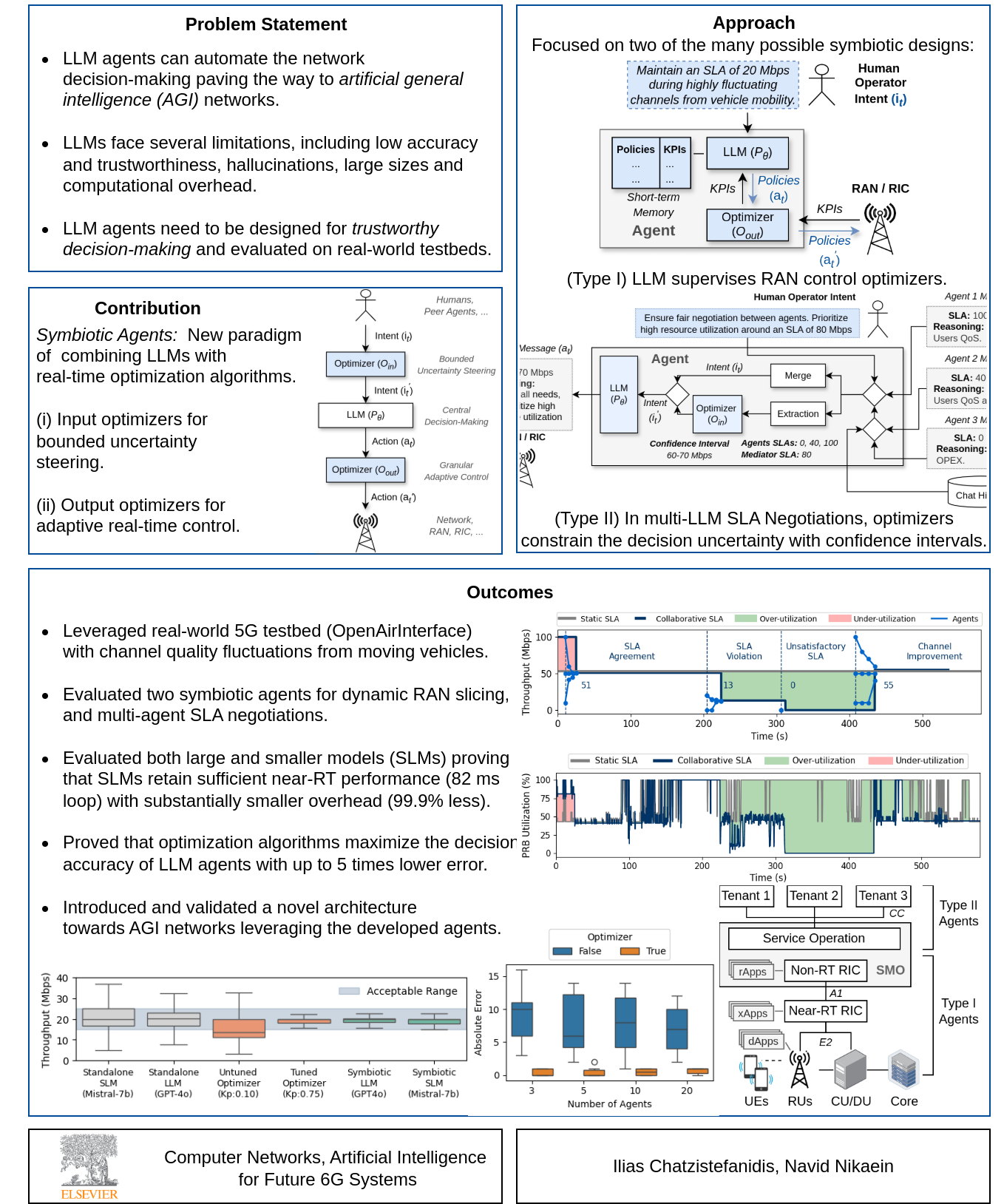}
\end{graphicalabstract}

\begin{highlights}
    \item Introduce a new agentic paradigm—\emph{symbiotic agents}—that pairs large language models with optimizers and examine it through the lens of trustworthy AI. 
    
    \item Implement two concrete agent designs for dynamic RAN control and multi-tenant SLA negotiations.

    \item Leveraged open-source real-world platforms, including OpenAirInterface (OAI) for the 5G user equipments (UEs), RAN and Core Network, and FlexRIC for the Radio Intelligent Controller (RIC).

    \item Utilized real-world datasets with RAN channel quality fluctuations from moving vehicles to emulate realistic mobility scenarios.

    \item Proved that optimization algorithms maximize the decision accuracy of LLM agents with up to $5$ times lower error.

    \item Evaluated both larger and smaller models (SLMs) proving that SLMs retain sufficient near-RT performance ($82$ ms loop) with substantially smaller GPU overhead (99.9\% less).

    \item Introduced and validated a novel LLM-driven architecture towards AGI networks leveraging the developed agents.   
\end{highlights}


\end{frontmatter}



\section{Introduction}

\label{sec:intro}

{By 2030 the number of fifth-generation (5G)—and early sixth-generation (6G)—subscriptions is forecast to exceed 6 billion \cite{gsma2025}.  
The radio environment will span $7$–$24\,$GHz spectrum sharing \cite{cui20236g}, integrated sensing and communication links, and digital-twin feedback loops \cite{wang2023road}.  
Such heterogeneity drives rapid spatio-temporal shifts in both user demand and channel condition, stressing every layer of resource-allocation logic.}

{Multi-tenant Radio Access Networks (RANs) let Mobile Network Operators (MNOs), Mobile Virtual Network Operators (MVNOs) and verticals co-habit common infrastructure \cite{samdanis2016network}.  
Intent-Based Networking (IBN) further hides low-level complexity behind declarative intents \cite{leivadeas2022survey}.  
The Open RAN (O-RAN) and newly formed AI-RAN Alliances embed \emph{artificial intelligence} (AI) into those abstractions as native control elements \cite{oran2025,airan2025}, yet today’s AI loops remain specialized and brittle.}

{\emph{Large Language Models} (LLMs) and their smaller counterparts, the \emph{Small Language Models} (SLMs), excel at high-level reasoning, fuelling visions of fully \emph{Artificial General Intelligence} (AGI)-driven networks \cite{bariah2024large,dou2023towards,saad2024artificial,bariah2023understanding}.  
However, they are probabilistic next-token predictors: they hallucinate facts \cite{huang2025survey, patil2024review}, break under \emph{out-of-distribution} (OOD) shifts \cite{yuan2023revisiting}, and lack formal safety guarantees.}

{Following the NIST AI Risk-Management Framework (AI-RMF) and ISO/IEC 42001, we define a trustworthy network agent as one that is \emph{robust}, \emph{interpretable}, \emph{secure}, \emph{fair} and \emph{governable} across its lifecycle \cite{nist_ai_rmf_2023, iso_iec_42001_2023}.  
LLMs alone satisfy only a subset of these attributes, and therefore more complex and robust agent architectures need to be explored.}

{We are the \emph{first} to formalize an agent architecture in which LLM reasoning is \emph{symbiotically} paired with deterministic optimization and present a full agent taxonomy with two concrete use cases:  
(i) in \textit{Type-I agents} for dynamic RAN control, the LLM interprets high-level intents and continually tunes the proportional gain~\(K_p\) of an underlying proportional (P-)controller, yielding certified, low-latency actions;  
(ii) in \textit{Type-II agents} for multi-agent service-level-agreement (SLA) negotiation, the LLM employs constraints that a gradient-descent optimizer produces to bound uncertainty and converge on fair resource allocations.  
Because guarantees are externalized to the optimizer, this synergy remains indispensable even as future LLMs improve, making the approach a necessity to bridge the gap towards AGI.}

{On a real-world 5G testbed that mirrors O-RAN/AI-RAN principles, we}  
\begin{enumerate}
    \item {\textit{define} the Symbiotic-Agent architecture under a unified trustworthiness lens;}  
    \item {\textit{instantiate} two agents—P-controlled slicing and gradient-bounded negotiation—achieving a $5{\times}$ reduction in decision error and up to 44\,\% spectrum savings;} 
    \item {\textit{benchmark} SLMs vs.\ LLMs, showing that prompt specialization shrink model size by up to 99.9\,\% without accuracy loss;}  
    \item {\textit{propose} an AGI-ready network architecture that encapsulates symbiotic agents as certified intelligence services.}  
\end{enumerate}

{The remainder of the paper formalizes the agent model (Section \ref{sec:symbiotic_agents}), details the testbed and evaluation (Sections \ref{ibn-arch}–\ref{eval}), shows an AGI-driven network use case (Section \ref{use_case}) and discusses open research directions toward fully AGI-driven networks (Section \ref{limitations}).}

\section{Related Work}
\label{relate_work}
\renewcommand{\arraystretch}{1.3} 

\begin{table*}[htbp]
\centering
\scriptsize
\begin{tabular}{@{}l p{0.1\linewidth} p{0.4\linewidth} p{0.4\linewidth}@{}}
\toprule
\textbf{Works} & \textbf{Category} & \textbf{Approach} & \textbf{Key findings} \\
\midrule

\rowcolor{lightgray}
\multirow{1}{*}{\cite{song2023self}} & Reinforcement learning & A LLM framework with a self-refinement mechanism for automated reward function design, where LLM can formulate an initial reward function based on natural language inputs. 
& LLM-designed reward functions can rival or even surpass manually designed reward functions in 9 robot control tasks. \\
\multirow{1}{*}{\cite{shinn2023reflexion}} & Agentic \newline Self Improvement & It proposed a novel framework to reinforce language agents through linguistic feedback. The agent verbally reflects on task feedback signals, maintaining the reflective text in an
episodic memory buffer to induce better decision-making. 
& The proposed framework achieves a 91\%
accuracy on the HumanEval coding
benchmark, surpassing the previous
state-of-the-art gpt-4 that achieves 80\%. \\
\rowcolor{lightgray}
\multirow{1}{*}{\cite{guo2023towards}} & Black-box \newline Optimizer & Evaluating the optimization capabilities of LLMs across
diverse tasks and data sizes, including gradient descent, hill-climbing, grid-search, and black-box optimization.
& 1) The LLM show strong optimization
capabilities; 2) LLMs perform well in
small-size samples; 3) They exhibit strong
performance in gradient-descent; 4) LLMs
are black-box optimizers. \\
\multirow{1}{*}{\cite{chen2024diagnosing}} & Convex \newline optimization & A natural language-based system that engages in interactive conversations about infeasible optimization models. It provides natural language descriptions of the optimization model itself, identifies potential sources of infeasibility, and
offers suggestions to make the model feasible. & The proposed system can assist both expert and non-expert users in improving their understanding of the optimization
models, enabling them to quickly identify the sources of infeasibility. \\
\rowcolor{lightgray}
\multirow{1}{*}{\cite{liu2025large}} & Heuristic \newline algorithms & Using general LLM serves as a black-box search operator for decomposition-based multi-objective evolutionary optimization in a zero-shot manner. & The LLM operator only learned from a few instances can have robust generalization performance on unseen problems with quite different patterns and settings. \\
\textbf{Ours} & Symbiotic Agents & A novel paradigm combining LLMs with optimization algorithms. Optimizers at the LLM's input-level provide bounded uncertainty steering for numerically precise tasks, whereas output-level optimizers supervised by the LLM enable adaptive real-time (RT) control.
& 
1) LLMs are meta-optimizers tuning parameters of control algorithms.
2) The uncertainty in LLM decision-making is efficiently bounded with confidence intervals decreasing the error up to $5$ times.
3) SLMs replace LLMs in near-RT tasks ($82~ms$ loops) maintaining accuracy and with up to $99.9$\% less graphics processing unit (GPU) overhead.
4) Novel AGI architecture is implemented with symbiotic agents.\\
\midrule
\end{tabular}
\caption{Comparison of the state-of-the-art on LLM-aided optimization with our approach.}
\label{tab:related-work}
\end{table*}

The telco industry explores LLMs to automate next-generation networks.
A collective effort from industry and academia charts the roadmap of large-scale AI adoption in telecom \cite{shahid2025large} explaining how Large Telecom Models (LTMs) could revolutionize the field. 
The literature proposes opportunities for applying LLMs on telecom divided into four categories \cite{10685369}, including (i) generation problems, (ii) classification problems, (iii) prediction problems and (iv) optimizing network performance. Generation problems include fine-tuning models on telecom domain question answering \cite{zou2024telecomgpt}, coding and troubleshooting, while the classification problems investigate network attack detection, and traffic classification \cite{shi2023bfcn}. The prediction problems, include traffic level forecasting \cite{service_aware1,service_aware2}, channel state estimation and user mobility predictions \cite{sousa2020survey, traffic_steer1, traffic_steer2}. 
Our work is positioned on the last category of network performance optimization, which includes LLM applications on real-time (RT) decision-making, such as resource allocation.

Table \ref{tab:related-work} summarizes the most prominent works on LLM-aided optimization, including our approach. 
An LLM framework with self-refinement mechanisms is developed \cite{song2023self} for automated reward function design of reinforcement learning (RL) algorithms. The results demonstrate that LLM rival manually designed reward functions in nine robot control tasks. 
In \cite{shinn2023reflexion} a novel framework reinforces language agents through linguistic feedback. The agent verbally reflects on task feedback signals, maintaining the reflective text in an episodic memory to induce better decision-making in the future. 
The authors in \cite{guo2023towards} evaluate the optimization capabilities of LLMs across
diverse tasks, including gradient descent, hill-climbing, grid-search, and black-box optimization, highlighting that LLMs are black-box optimizers.
In \cite{chen2024diagnosing} authors work on a natural language-based system that engages in interactive conversations about infeasible optimization models. 
In \cite{liu2025large} an LLM serves as a black-box search operator for decomposition-based multi-objective evolutionary optimization in a zero-shot manner. 
These works illustrate the abilities of LLMs as black-box optimizers as well as handling external optimization techniques. 

Following these motivations, some works start applying LLMs in real network systems. In \cite{zou2023wireless} the authors design on-device LLMs, where multi-agent LLMs are solving network tasks in a game theoretic manner. 
In \cite{ameur2024leveraging} a framework is proposed that leverages LLMs and prompt engineering techniques to elucidate RL algorithms’ decision-making showcasing improvements in comprehensibility for network slicing. 
In \cite{wu2025llm} they propose an intelligent LLM agent prompting to dynamically optimize resource allocation of network slices. 
An LLM-centric Intent Life-Cycle (LC) management architecture \cite{mekrache2024intent} is designed to manage network services using natural language. 
\textsc{Maestro} \cite{maestro} is the first work to propose an LLM-based business plane for collaborative multi-tenant decision-making on a real testbed showing the vision towards AGI networks. 
\textsc{Agoran} \cite{chatzistefanidis2025agoran} deploys a digital agora on top of 6G networks inspired by ancient Greek agora following a tripartite architecture. It is the first work to formally utilize and scale the \emph{symbiotic paradigm} integrating a multi-objective optimizer (NSGA-II) to provide a Pareto front of near optimal SLA offers for multi-stakeholder LLM bargaining.
MX-AI \cite{chatzistefanidis2025mx} connects a multi-agent graph into the R1 interface of Open RAN networks capable of intent-based observability and control actions; thus creating an open platform to accelerate future research towards agentic AI-driven RANs.

These works show the great potential of LLMs on network tasks but lack providing comprehensive analysis on the trustworthiness of the LLM decision-making for realistic systems.
Importantly, there is a major need for guarantees and uncertainty bounds for improving the agents accuracy. 
Also, we identify a gap on highly variable channels in low-latency sub-millisecond timing loops. 
Further, scaling down the size and overhead of models by utilizing SLMs is essential to minimize the cost at sustainable levels and need to be proven achievable in real network systems. 
These reasons lead us to propose a novel agentic paradigm to improve the LLM towards trustworthy decision-making by combining it with optimization algorithms. 
Optimizers at the LLM's input-level provide bounded uncertainty steering with confidence intervals for numerically precise tasks, whereas output-level optimizers are supervised by the LLM for adaptive real-time control.
We work on real systems, where resources and intents are varying, using emulated channel fluctuations from moving vehicles, evaluating both large and smaller models. 
We focus on two distinct cases of RAN slicing and multi-tenant SLA negotiations. 
Based on our results, we propose a novel AGI-driven network architecture.

\section{Symbiotic Agents: Architecture and Trustworthiness Lens}
\label{sec:symbiotic_agents}

{Autonomous network agents must satisfy a \emph{comprehensive} trustworthiness profile that includes and extends well beyond decision-making accuracy: \emph{robustness} to channel and workload drift, \emph{interpretability} for human audit, \emph{security} against adversarial manipulation, \emph{fairness} across tenants, and \emph{governance} over the agent’s entire lifecycle \cite{nist_ai_rmf_2023, iso_iec_42001_2023}.}

{LLMs excel at high-level reasoning, transparent chain-of-thought logging, and natural-language justification, yet they provide no formal guarantees on numeric error or worst-case latency. Conversely, control-theoretic and optimization routines deliver deterministic performance bounds but lack semantic understanding. \textit{Symbiotic Agents fuse these complementary strengths into a single loop, thereby covering all five trustworthiness pillars.} Table~\ref{tab:trustworthiness_map} summarizes how each pillar is realized in our architecture.}

\begin{table*}
    \centering
    \scriptsize
    \begin{tabular}{@{}p{2.6cm}p{10cm}@{}}
        \toprule
        \textbf{Pillar} & \textbf{Mechanism in Symbiotic Agents} \\
        \midrule
        {Robustness} &
        {Input-side optimizer guards against out-of-distribution SLA bids; Output-side optimizer supplies real-time deterministic error bound $\varepsilon$ with the Memory-based LLM meta-control adapting to channel variability.} \\
        {Interpretability} &
        {LLMs produce explicit chain-of-thought and structured artifacts
        (JSON SLA decisions and rationales);
        all traces logged by logs $\mathcal{L}$ for audit.} \\
        {Security} &
        {Numeric guard-rails and KPI triggers prevent adversarial or hallucinated actions from propagating to the real-time loop.} \\
        {Fairness} &
        {Tenant Mediator utility \(U_0\) plus SLA confidence-interval prompts enforce symmetric treatment of all tenants.} \\
        {Governance} &
        {Logs $\mathcal{L}$ and KPIs provide lifecycle logging;
        compliance monitoring of the KPIs and observance of deviations could trigger rollback.} \\
        
        {LLM$\rightarrow$AGI gap} & {Input-side optimizer bounds uncertain inputs; output-side optimizer attaches deterministic action certificates; logger $\mathcal{L}$ keeps an auditable trace. Three properties unattainable through LLM scale alone.} \\

        \bottomrule
    \end{tabular}
    \caption{{Mechanisms by which Symbiotic Agents (i) realize the five trustworthiness pillars and (ii) provide a durable bridge from current LLM capabilities to AGI-grade guarantees.}}
    \label{tab:trustworthiness_map}
\end{table*}

\subsection{{Formal Definition}}
\label{sec:formal_def}

\begin{definition}[Symbiotic Agent]
\label{def:symb_agent}
{A \emph{Symbiotic Agent} is the quintuple
$\mathcal{A}=\langle\mathcal{E},\,
                 \mathcal{P}_{\theta},\,
                 \mathcal{O}_{\text{in}},\,
                 \mathcal{O}_{\text{out}},\,
                 \mathcal{L}\rangle$ where}
\begin{enumerate}
\item {$\mathcal{E}$ is the partially observable network environment (KPIs, channel state, tenant intents, negotiation messages);}
\item {$\mathcal{P}_{\theta}$ is an LLM that maps a contextual
      prompt~$i_t$ to a structured action artifact~$a_t$ ($Kp$ hyper-parameters, SLA bids);}
\item {$\mathcal{O}_{\text{in}}$ (\emph{input-side optimizer})
      optionally pre-processes $i_t$ to a bounded $i_t'$ supplying numeric guard-rails
      before reasoning (e.g.\ a confidence interval for SLA values);}
\item {$\mathcal{O}_{\text{out}}$ (\emph{output-side optimizer})   
      optionally converts $a_t$ into a granular  action $a_t'$ with a provable error bound
      $\varepsilon$ (e.g.\ P-control on Physical Resource Blocks (PRB) allocation);}
\item {$\mathcal{L}$ logs the internal trace for audit and feeds
      compliance signals back to $\mathcal{P}_{\theta}$.}
\end{enumerate}

{The loop $i_t\xrightarrow{\mathcal{O}_{\text{in}}}
        i_t'\xrightarrow{\mathcal{P}_{\theta}}
        a_t\xrightarrow{\mathcal{O}_{\text{out}}}a_t'$
executes at two time scales: sub-second LLM updates $(\mathcal{P}_{\theta})$ and
sub-millisecond numeric control $\mathcal{O}_{\text{in}}, \mathcal{O}_{\text{out}}$, ensuring both semantic richness and
hard-real-time guarantees.}
\end{definition}

\begin{figure}
    \centering
    \includegraphics[width=0.35\linewidth]{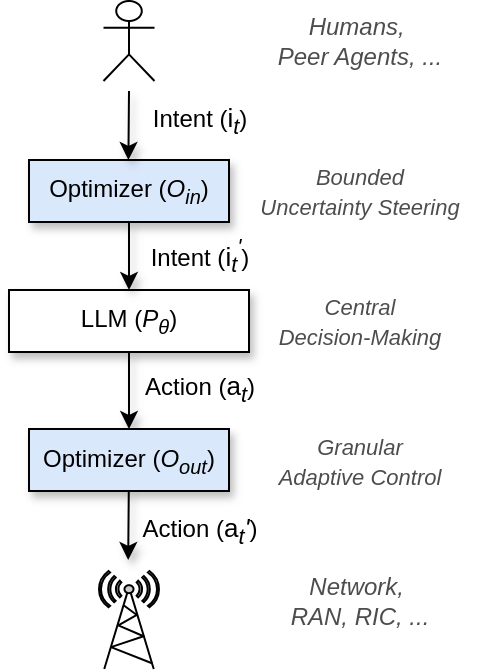}
    \caption{LLM \textsc{Symbiosis} Paradigm: Input-level optimizers provide bounded uncertainty steering for numerically precise tasks, whereas output-level optimizers enable adaptive real-time control actions.}
    \label{fig:symbiosis-concept}
\end{figure}

{Figure~\ref{fig:symbiosis-concept} visualizes this layered control
loop, showing how the input-side and output-side optimizers flank the
LLM to deliver both bounded uncertainty and real-time numeric
certification.}
{The authors believe that this symbiotic design is indispensable for trustworthy agents towards AGI. Even when future LLMs improve, next-token sampling remains stochastic and cannot yield deterministic error bounds.  Therefore
$\mathcal{O}_{\text{in}}$ and/or $\mathcal{O}_{\text{out}}$
remain indispensable to close the gap towards AGI-grade decision-making and trustworthiness.}

\subsection{Type I Agent: Granular Adaptive RAN Control}
\label{type1_design}

\begin{figure}
    \centering
    \includegraphics[width=0.55\linewidth]{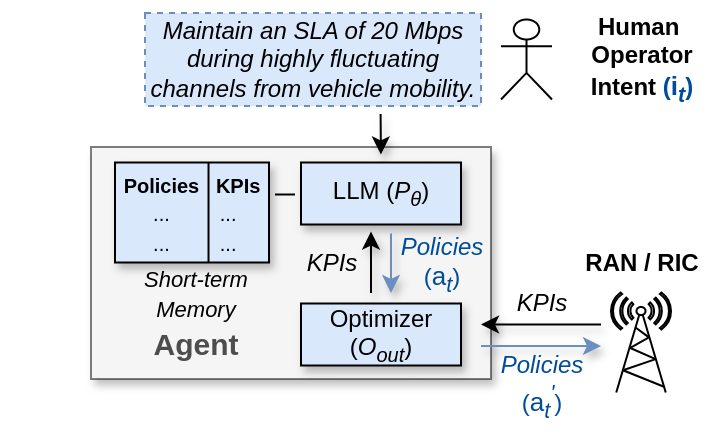}
    \caption{Type I Symbiotic Agent for RAN Control. An LLM {($\mathcal{P}_{\theta}$)} translate operator intents {$(i_t)$} to enforceable actions {($a_t$)} by deploying and supervising RAN control optimizers {as $\mathcal{O}_{\text{out}}$, which ultimately control RAN resources, such as PRBs ($a_t'$)}. The LLM works as a black-box meta-optimizer, continuously tuning the hyperparameters of the control algorithms adapting to variable channels.}
    \label{fig:type1_sym}
\end{figure}

Figure \ref{fig:type1_sym} shows a Type I symbiotic agent designed for closed-loop RAN {control}. It receives the operator intent and deploys a control optimization algorithm {$\mathcal{O}_{\text{out}}$} to allocate real-time resources in the RAN. Typical control algorithms include rule-based, reinforcement learning or control-theoretic methods. These algorithms are generally accurate but rely heavily on their hyper-parameter tuning, especially in highly variable networks. This is because unpredicted interference and channel fluctuations disrupt the algorithm's efficiency leading to oscillations and need for re-exploration or tuning. In our experiments we employ a control-theoretic approach, following the widely adopted Proportional–Integral–Derivative (PID) Controller \cite{visioli2006practical}, {with section \ref{sssec:pid-rationale} explaining the rationale behind this choice}. 
The algorithm is robust to channel variability bringing the control variable to the desired state, without the need of training data. It is heavily influenced by its parameter selection. We use the simplified version of the proportional Control (P-control) algorithm, as it is proved sufficient in our experiments. Specifically, P-control is a type of linear feedback control system, in which a correction is applied to the controlled variable-here RAN physical resource blocks (PRBs)- and the size of it is proportional to the difference between the intent and current state.
For instance, a throughput (Tp) intent is enforced by controlling the PRB utilization capacity of a RAN slice ranging from $0$ to $100$\%. Thus:

\begin{equation}
    PRB^{new} = K_pe(t) + PRB^{current}, 
\end{equation}   
where $K_p$ is the proportional gain, $e(t)$ is the instantaneous process error at time $t$:
\begin{equation}
    e(t)=Tp^{Intent}-Tp^{current}
\end{equation}

After deploying the control algorithm, the LLM works continuously as a meta-optimizer to fine-tune the algorithms hyper-parameter (here $K_p$). 
As shown in Figure \ref{fig:type1_sym}, it reads key performance indicators (KPIs) of the algorithm performance, checks past action sets from a short-term memory and send updated policies for its configuration.
In our experiments, as a KPI the LLM reads the P-control's average number of iterations to converge in the last couple of time steps, {as explained in section \ref{sssec:granular}}.
If it is larger than a threshold, here two iterations, the LLM updates the $Kp$ value exploring a better configuration.
This way, it adapts to the variable channels ensuring algorithmic convergence to a desired performance (e.g. only two iterations to find optimal PRB).
The short-term memory collects knowledge from past actions with the LLM self-improving in an agentic manner and reducing the repetition of errors.
This symbiotic design positions the LLM at an appropriate abstraction level, operating in near-real-time (near-RT) loops ($\geq10$ ms), while the control algorithm works in real-time sub-millisecond ones ($\leq1$ ms).

\subsubsection{{Granular Adaptive Control:  How the LLM Steers the P–Controller}}

\label{sssec:granular}
\begin{figure}
    \centering
    \includegraphics[width=0.6\linewidth]{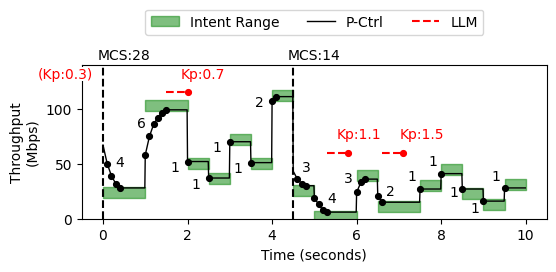}
    \caption{{Timeline of a Type I Symbiotic Agent executing \emph{granular adaptive control}. The black curve is the throughput obtained by the \emph{P-Control} loop ($\leq10ms$); the green band marks the operator’s intent range. Digits next to the black markers indicate how many \emph{P-Control} iterations (\(\iota\)) were required to meet each intent. The channel quality changes from MCS 28 to MCS 14 at the dashed vertical line. After every 3–5 enforcements the LLM computes the average \(\overline{\iota}\); whenever \(\overline{\iota}>2\) it updates the proportional gain \(K_p\) (red dashed markers), raising it from 0.3 to 0.7, 1.1, and finally 1.5. These meta-control interventions—issued only once \(\overline{\iota}>2\)—restore single-iteration convergence while keeping the fast, numerically precise \emph{P-Control} loop entirely intact.}}
    \label{fig:type1_loops}
\end{figure}

{Figure~\ref{fig:type1_loops} zooms in on the internal feedback structure of a \textit{Type I Symbiotic Agent}. The low-level P-Control is the point of contact with the RAN scheduler; it executes in $10^{-4}$ milliseconds (ms) and is therefore able to react to every $5$ ms change in measured throughput}\footnote{{The figure shows the throughput \emph{after} the $5$ ms air-interface reaction time on our testbed.}}. {The LLM operates as an asynchronous \emph{black-box meta-optimizer} and is invoked only when P-Control shows signs of sluggishness. This two-time-scale design keeps the fast loop entirely numerical while delegating rare but expensive reasoning to a slower cognitive loop.}

\noindent\textit{{What the LLM sees.}}  
{After every \textit{cluster} of $N\!=\!3$--$5$ intent enforcements by {P-Control}, we compute a single key performance indicator (KPI):}

\begin{equation}
    {\overline{\iota}(t)=\frac{1}{N}\sum_{j=1}^{N} \iota_j}
\end{equation}

{where $\iota_j$ is the number of control iterations \emph{P-Control} needed to meet intent~$j$.  
This average iteration count is the \emph{KPI} passed to the LLM
together with the last ten
\(\langle K_p,\overline{\iota}\rangle\) pairs stored in a small
short-term memory.  
Keeping the memory to ten entries bounds prompt length and limits the
SLM’s inference time to 82 ms (or 1000 ms for the full LLM), which makes it suitable for near-RT loops ($10~ms \leq x \leq 1~sec$)}

{\noindent\textit{When it is activated.}  
If \(\overline{\iota}(t)\le\tau\) with
\(\tau=2\), the system is already converging in at most two control
iterations and no action is taken.  
Otherwise the LLM is triggered.}

{\noindent\textit{How it updates \(K_p\).}  
The prompt contains natural-language instructions that implement the
following heuristic:}

\begin{enumerate}
    \item {We rank the past memory actions \(\langle K_p,\overline{\iota}\rangle\) by recency in a time-series manner.}
    \item {We prompt the LLM to observe whether
      \(\overline{\iota}\) has \emph{decreased} as $K_p$ was increased. If yes, continue in the same direction; otherwise reverse.}
    \item {We prompt the LLM to freely choose the new configuration \(K_p^{\text{new}},\) with $K_p^{\text{new}}\in(0,~\infty)$.}
\end{enumerate}

{The new $K_p$ gain is streamed back to the \emph{P-Control}, while \emph{P-Control} keeps running, making the large-model latency invisible to the real-time loop.}

{\noindent\textit{Illustrative trace (Figure~\ref{fig:type1_loops}).}  
Starting with \(K_p{=}0.3\) under a favorable channel with RAN modulation and coding scheme (MCS) of 28, \emph{P-Control} needs 4–6 iterations to hit the intent
(\(\overline{\iota}=5>\tau\)); the LLM therefore raises the gain to
\(0.7\), after which convergence occurs in 1–2 iterations.  
When mobility degrades the channel to MCS 14 the same gain is again too
small (\(\overline{\iota}=3.5\)), prompting two further LLM
updates—first to \(1.1\) and finally to \(1.5\)—until every subsequent
intent is met in a single iteration \(\overline{\iota}=1.0\).  
Throughout the run the achieved throughput (\emph{P-Control} black curve) remains inside the green intent band, showing that the LLM meta-controller adjust flexibly the performance of \emph{P-Control} across channel variability.}

{\noindent\textit{Ablation on memory length.}  
Removing the memory forces the LLM to explore blindly and
increases the number of LLM invocations as we discuss in evaluation section \ref{memory-eval}.
Expanding the memory beyond ten entries yields no high additional accuracy but pushes LLM latency above the $1$ sec near-RT budget.  
A ten-entry window therefore strikes the best balance between knowledge retention and computational cost.}

\subsubsection{{Why a PID Inner Loop?  Comparison With Traditional Controllers}}
\label{sssec:pid-rationale}

\begin{figure}[t]
    \centering
    \begin{subfigure}[b]{0.45\textwidth}
        \centering
        \includegraphics[width=0.99\linewidth]{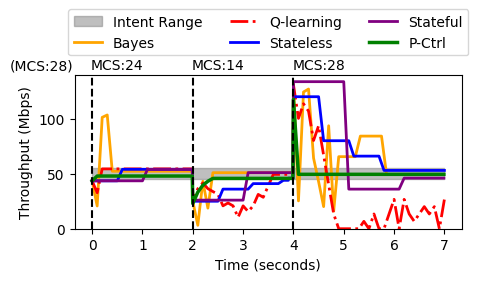}
        \caption{{Full comparison: traditional controllers and direct–LLM baselines.}}
        \label{fig:type1_full_cmp}
    \end{subfigure}
    \hfill
    \begin{subfigure}[b]{0.52\textwidth}
        \centering
        \includegraphics[width=0.99\linewidth]{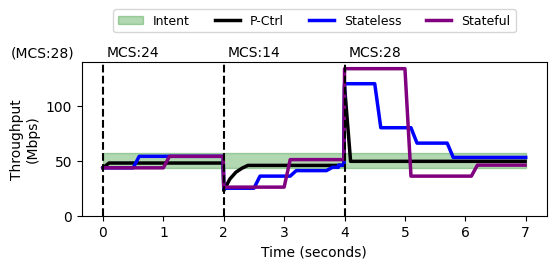}
        \caption{{Zoom on \textit{P-Cotrol} compared with the standalone LLMs.}}
        \label{fig:type1_sym_cmp}
    \end{subfigure}
    \caption{{Throughput under a fluctuating channel (MCS $28\,\rightarrow\,24\,\rightarrow\,14\,\rightarrow\,28$).
             The green band marks the intent range.}}
    \label{fig:type1_alg_cmp}
\end{figure}

{Figure~\ref{fig:type1_alg_cmp} benchmarks five candidate controllers under the same variable–channel trace, including Bayesian optimization and reinforcement learning (RL) methods. The curves motivate our choice of a \emph{Proportional controller} as the inner loop that the LLM meta-optimizer tunes.}
\textit{{Bayesian optimization}}{ (orange, Figure\,\ref{fig:type1_full_cmp})%
      explores aggressively; the resulting large steps overshoot the
      intent and create visible oscillations.  Moreover, each probe
      requires a posterior update, which makes the algorithm slower in time.}
 {\textit{Q-learning}, a widely used RL technique, (red dashed) converges when the channel is
      stationary (MCS 28–24) but must \emph{re-explore} after the drop to MCS\,14, producing a performance dip—unacceptable for real-time control where channel conditions can change every frame.}
{Stateless (blue) and stateful (purple) standalone LLMs track the intent without oscillation, yet their $100$–$1000$ ms inference latency makes them unsuitable for a sub-ms scheduler loop.  Figure~\ref{fig:type1_sym_cmp}
confirms more clearly that both LLM variants, although stable and accurate, they lag behind the numeric \emph{P-Control} base
line (black).}

{The following arguments consolidate our selection of P-Control for the Type I symbiosis with a hypervisor LLM.} {(i) \textit{Robust\,\&\,lightweight.} For a first-order plant (resource-block
            allocation to throughput) a proportional controller is the
            minimal structure that guarantees closed-loop stability~\cite{astrom_pid_2006}. It is
      model-free, requires \emph{no} training, and executes in
      \(10^{-4}\) ms on our radio, leaving ample headroom beneath the 5 ms air-interface delay.}
{\textit{(ii) Single hyper-parameter.}  Its only sensitivity is the gain
      \(K_p\).  This makes it ideal for a symbiosis in which an LLM can
      focus on \emph{one} continuous tuning knob.}
{\textit{(iii) Natural division of labor.}  The inner loop (\emph{P-Control}) delivers
      sub-ms numerical precision; the LLM supplies zero-touch
      \emph{granular adaptive control}, adjusting \(K_p\) whenever the
      KPI \(\overline{\iota}\) signals sluggish convergence
      (section \ref{sssec:granular}).  The resulting hybrid keeps the best of
      both worlds: the speed of control theory and the flexibility of
      large-scale reasoning.}

\subsection{Type II Symbiosis for Multi-Agent SLA Negotiations}
\label{type2_agent_section}

\begin{figure}
    \centering
    \includegraphics[width=0.5\linewidth]{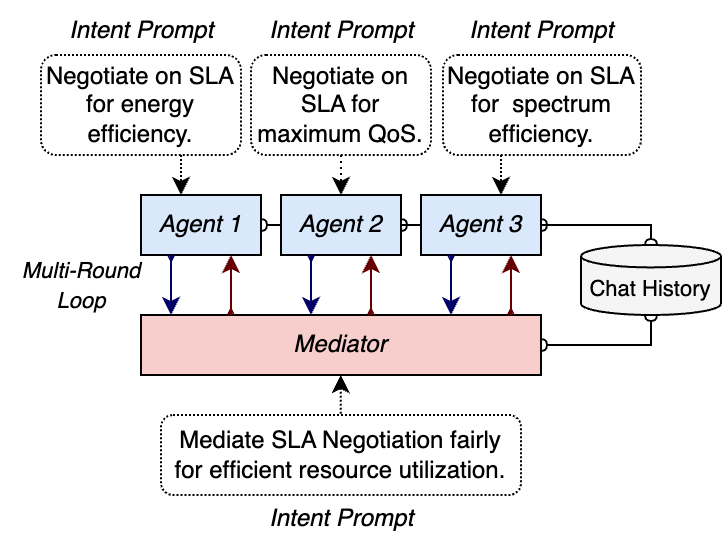}
    \caption{Multi-Tenant Negotiation Topology where multiple agents (tenants) negotiate towards an SLA consensus guided by a network mediator (MNO). The multi-round negotiations take into account the collective objectives to converge on a Pareto-optimal SLA.}
    \label{fig:mas-topo}
\end{figure}

A central aspect of multi-agent systems (MAS) is negotiation, often modeled as a distributed optimization problem in which multiple agents must arrive at a consensus \cite{scutari2013decomposition, wang2020distributed}. From a game-theoretic perspective, an optimal consensus is a Pareto-efficient Nash Equilibrium (NE) that balances both individual and collective objectives.
Nash Equilibria represent stable consensus points where no agent can unilaterally improve its outcome. However, when LLM agents engage in negotiations, issues such as hallucinations or non-cooperative (overly greedy) behaviors can disrupt the process, shifting the NE away from Pareto efficiency. Figure \ref{fig:mas-topo} illustrates a multi-tenant MAS topology built around a central mediator, typically operated by the MNO, which guides negotiations toward fair, cooperative, and Pareto-efficient equilibria.

\begin{figure*}
    \centering
    \centering
    \includegraphics[width=0.99\textwidth]{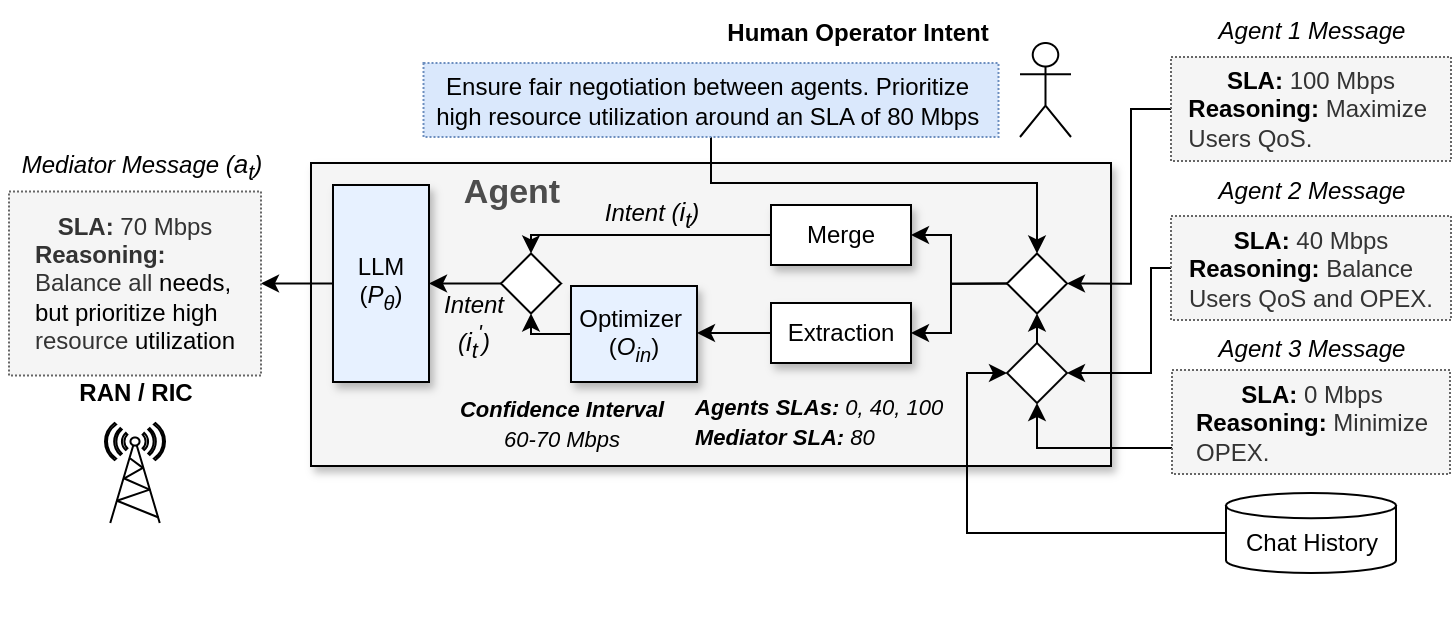}
    \caption{Type II Symbiotic Agent for SLA Negotiations: Intent Prompt is merged with the messages of the agents in the current negotiation round along with the chat history {to unified intent ($i_t$)} to be provided to the LLM. A side-car optimization algorithm {($\mathcal{O}_{\text{in}}$)} extracts the proposed SLA values calculating and {adding an SLA confidence interval to the final intent ($i_t'$)} to constraint the uncertainty and steer the {LLM negotiations ($\mathcal{P}_{\theta}$)} towards a Pareto-optimal solution.}
    \label{fig:type2_sym}
\end{figure*}

Figure \ref{fig:type2_sym} illustrates a Type II agent designed for multi-tenant SLA negotiations. 
The agents negotiate on finding the Pareto-optimal SLA, using a standardized structure, without loss of generality, consisting of two parts: (a) the proposed SLA and (b) the decision reasoning in natural text. This way, the SLA proposal (e.g. RAN throughput) is supported by detailed reasoning for LLM decision-making interpretability, which considers individual and collective objectives.
Each agent merges the human operator intent (from tenants or MNO), the messages of other agents, and the negotiation chat history to a unified intent ($i_t$) to take a decision.

\textit{Optimization Algorithm.}
\label{optimizer_unit}
A side-car optimizer {($\mathcal{O}_{\text{in}}$)}, extracts the numerical SLA values (e.g. throughput) and runs a gradient-descent (GD) algorithm modeling the specific topology set up. 
Then, it calculates a confidence SLA interval (e.g. $60$-$70$ Mbps), where the Pareto-optimal value lies with a degree of confidence (e.g. 95\%). 
\textit{Utility Functions.}
The algorithm models the topology, with each agent having an individual utility function:
\begin{equation}
    U_i(x_i) = - \alpha_i (x_i - d_i)^2,
\end{equation}

where \(d_i\) is the agent's desired SLA and \(\alpha_i > 0\) measures sensitivity to deviations. 
The mediator maintains a global utility function:
\begin{equation}
    U_0(x_1,\dots,x_n) = - \gamma \sum_{i=1}^{n} (x_i - \bar{x})^2 \;-\; \beta \bigl(\bar{x} - x_{\text{target}}\bigr)^2,
\end{equation}

where \(\bar{x} = \tfrac{1}{n}\sum_{i=1}^n x_i\), \(x_{\text{target}}\) is the mediator's preferred SLA, and \(\gamma,\beta > 0\) are weighting factors for consensus and alignment to \(x_{\text{target}}\), respectively.

\textit{Optimization Formulation.}
We combine these utilities into a single objective to maximize:
\begin{equation}
    \max_{x_1,\dots,x_n} \sum_{i=1}^{n} U_i(x_i) + \lambda \, U_0(x_1,\dots,x_n),
\end{equation}

where \(\lambda\) balances individual vs. global objectives. Since each \(U_i\) and \(U_0\) is concave in the \(\{x_i\}\), their sum is also concave, making this optimization well-behaved under mild assumptions.

\textit{Gradient-Based Update.}
We iteratively apply gradient descent to the negative of our objective. Each agent \(i\) updates \(x_i\) by:
\begin{equation}
    x_i^{(k+1)} = x_i^{(k)} - \eta \Bigl[ 2\,\alpha_i \bigl(x_i^{(k)} - d_i\bigr) \;+\; 2\,\gamma \bigl(x_i^{(k)} - \bar{x}^{(k)}\bigr) \;+\; \beta \bigl(\bar{x}^{(k)} - x_{\text{target}}\bigr) \Bigr],
    \label{iterative_algo}
\end{equation}

where \(\eta > 0\) is the learning rate, and we clamp \(x_i\) to \([0,100]\) as the SLA range. If the system satisfies
\(\max_i \lvert x_i^{(k)} - \bar{x}^{(k)}\rvert < \epsilon\),
we declare consensus convergence and return the average \(\bar{x}^{(k)}\).

\textit{Empirical Validation and Confidence Intervals.}
To assess the robustness of the solution, we perform \(R\) independent optimization runs, each with perturbed initial demand values \(\{d_i\}\). In each run \(r = 1, \dots, R\), we obtain a final consensus value \(\hat{x}^*_r\). From these, we compute the sample mean \(\overline{x^*}\) and standard deviation \(s\), forming a 95\% confidence interval:
\begin{equation}
    \overline{x^*} \;\pm\; 1.96 \,\frac{s}{\sqrt{R}}.
    \label{conf_interv}
\end{equation}

This interval reflects the variability in consensus outcomes due to uncertainty in initial conditions. It serves as a statistical estimate of the true optimal SLA, and is used to guide the LLM agents. It promotes convergence to values close to the Pareto-optimum with high reliability and thus improving the decision robustness in out-of-distribution bids.

This interval is appended to the unified intent $i_t$ forming a bounded intent {($i_t'$)}. This symbiotic synergy provides to the LLM the appropriate context, bounding the SLA uncertainty and allowing it to consolidate an optimal final decision.

\subsubsection{{Uncertainty Bounding:  Confidence Intervals Interaction with LLMs.}}
\label{sssec:llm-opt-type2}

\begin{figure}
    \centering
    \includegraphics[width=0.7\linewidth]{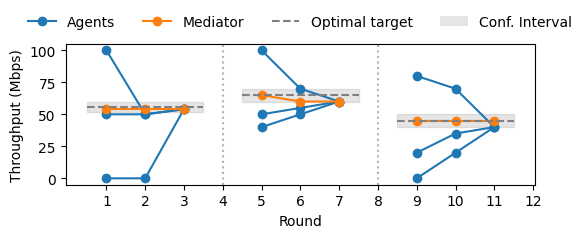}
    \caption{{Three negotiation Games using a pre-negotiation pipeline that bounds numerical uncertainty for all
    LLM agents.  A light-weight optimizer computes a confidence interval
    $\mathcal{C}=[L,U]$ for the Pareto–optimal SLA and injects it into the
    prompt of every agent (including the mediator) before the first dialogue
    round.  Subsequent bids must stay within $\mathcal{C}$; natural-language
    argumentation is unconstrained.  The game ends when the spread of the bids
    shrinks below an $\epsilon$-wide band.}}
    \label{fig:conf_intrvl_mechanism}
\end{figure}

{Figure \ref{fig:conf_intrvl_mechanism} shows how the single, millisecond-scale optimization step ($\mathcal{O}_{\text{in}}$) \emph{pre-shapes} an entire SLA-negotiation game.}
{Immediately after the tenants and the MNO-mediator submit their intent prompts, the initial numeric claims are extracted, producing the vector $\mathbf{x}^{(0)}=[x^{(0)}_1,\dots,x^{(0)}_n]$.}
{The side-car optimizer runs the gradient scheme of
Eq.~\eqref{iterative_algo} for $R=100$ independent restarts, each seeded with a jittered copy of $\mathbf{x}^{(0)}$.  From the resulting sample distribution it computes a mean $\overline{x^*}$ and a 95\,\% confidence interval $\mathcal{C}=[L,U]$ Eq.~\eqref{conf_interv}.  The whole batch completes in \(<\!1\) ms on a commodity CPU.}
{The interval $\mathcal{C}$ is appended to the prompt of \emph{every} LLM agent, prefixed by a short instruction:}

\begin{flushleft}
\fbox{\parbox{0.96\linewidth}{
\small
\texttt{{Numerical guard-rail: Offer an SLA strictly within $[L,U]$.  If you propose a value outside this interval, justify the trade-off explicitly.}}
}}
\end{flushleft}

{Agents now exchange natural-language proposals.  Any numeric bid must lie inside $[L,U]$; otherwise the message must include a clear justification.  This rule throttles overly greedy or hallucinated bids while leaving higher-level reasoning—and therefore fairness, persuasion, or strategic concessions—fully under LLM control. In our experiments the parties converge in two to five rounds ($\sim$10-48 s wall-clock time).}

\noindent\textit{{Remark.}}  
{The interval can be recomputed after every round—yielding a tighter bound at the cost of extra CPU time—but a single pre-negotiation pass is sufficient for all scenarios tested in this paper.}
{This uncertainty-bounding mechanism ensures that numerically precise tasks stay well-behaved, while the LLMs remain responsible for the rich, explainable reasoning that ultimately convinces all parties to accept a common SLA.}

\subsubsection{{Why a Gradient Descent Side-Car Optimizer?}}
\label{sssec:gd-rationale}

\begin{figure}
    \begin{subfigure}[b]{0.49\linewidth}
        \centering
        \includegraphics[width=0.7\linewidth]{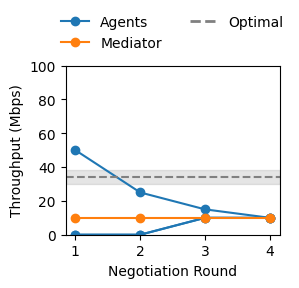}
        \caption{{Standalone Negotiations without SLA Confidence Interval.}}
        \label{fig:no-conf-int}
    \end{subfigure}
    \hfill
    \begin{subfigure}[b]{0.49\linewidth}
        \centering
        \includegraphics[width=0.7\linewidth]{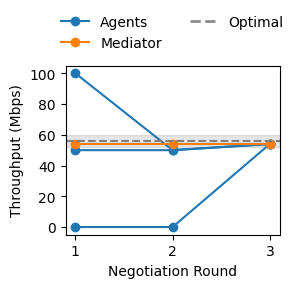}
        \caption{{Negotiations supported by the SLA Confidence Interval.}}
        \label{fig:with-ci}
    \end{subfigure}
   
    \caption{{Multi-Agent negotiations employing SLA confidence intervals for reducing the agents' decision uncertainty converging towards a Pareto-optimal solution.}}
     \label{confidence_interval}
\end{figure}

{Figure~\ref{confidence_interval} contrasts two negotiation games:
(a)~stand-alone LLM agents and
(b)~LLM agents steered by the confidence interval $\mathcal{C}$ computed by our side-car optimizer.  
Without $\mathcal{C}$ (Figure\,\ref{fig:no-conf-int}) the agents converge cooperatively yet stabilize \emph{outside} the Pareto-optimal target. With $\mathcal{C}$ (Figure\,\ref{fig:with-ci}) every bid remains inside the shaded band and the process terminates near the optimum.}

{Below we explain why a \emph{simple GD} solver is the most suitable engine for producing that bound.}
{(i) \textit{Problem structure.}}
{The joint objective in Eq.\,\eqref{iterative_algo} is the sum of concave utilities
$U_i(\cdot)$ and a concave mediator regularizer $U_0(\cdot)$; therefore the optimization landscape is \emph{strictly concave} and admits a \emph{unique} maximizer \cite{boyd_convex_2004}.  
For such problems deterministic first-order GD converges
geometrically~\cite{bertsekas_nonlinear_1999} and does not require
second-order information or black-box sampling.}
{(ii) \textit{Predictable and lightweight.}}
{The optimizer touches a single scalar variable per agent, so a full
GD pass costs $\mathcal{O}(n)$ arithmetic operations and completes in
$<\!1$\,ms on a commodity CPU (section \ref{sssec:type2-bench}).  
This deterministic runtime is crucial: the bound must be available
\emph{before} the first LLM round and must not enlarge the
near-RT budget dictated by language-model
inference ($\ge\!100$\,ms).}
{{(iii)\textit{Direct link to statistical uncertainty.}}}
{Running $R=100$ independently jittered restarts yields an empirical
distribution $\{\hat{x}^*_r\}$ from which we derive the 95\,\%
confidence interval
$\mathcal{C}$.  
GD’s negligible run-time makes this Monte-Carlo style bootstrap
feasible at each game initialization}\footnote{{Recomputing the interval
every round is possible (see remark in section\,\ref{sssec:llm-opt-type2})
but was unnecessary in all test cases, and could be explored in the future.}}.
{{(iv)\textit{Compatibility with LLM prompting.}}}
{The optimizer returns only \emph{two} numbers, $L$ and $U$, which fit
cleanly into a single guard-rail instruction
(\texttt{``Offer an SLA strictly within [L,U]''}).  
This keeps the prompt length constant and avoids the chain-of-thought
leakage that larger numeric payloads can cause in LLMs.}

{\textit{Future Notes.} Although first-order GD best fits the present study’s real-time and
single-dimension setting, we recognize that more sophisticated
optimizers could unlock additional capabilities:}
{\textit{(i) Reinforcement Learning with Human Feedback (RLHF).}
      Policy-gradient methods such as proximal policy optimization (PPO) can learn nuanced,
      preference-aligned bargaining strategies and might outperform
      fixed heuristics when the negotiation objective spans multiple,
      non-convex dimensions (e.g.\ QoS \emph{and} carbon footprint).
      Their ability to incorporate human reward signals could make the
      mediator more transparent and user-controllable.}  
{\textit{(ii) Bayesian Optimization.}
      For future scenarios involving many coupled SLA variables,
      Bayesian search would provide principled exploration – exploitation
      trade-offs and native uncertainty quantification.  Gaussian-process
      posteriors could also feed richer priors back into the LLM
      prompt, beyond the simple interval used here.}
{Both approaches, however, come with practical costs—substantial sample requirements for RLHF and cubic-time kernel updates for Bayesian optimization—that could exceed the 2–4 round budget and sub-second delay targets of our current prototype. Investigating these richer, but more expensive, optimization layers therefore remains a promising avenue for future work once stricter latency constraints are relaxed or more compute is available at the network edge.}

\textit{Integrating Advanced Reinforcement Learning.}
While deterministic optimizers such as GD and P‑controllers are ideal for the strict sub‑millisecond budgets of our inner‑loop agents, more expressive reinforcement‑learning (RL) methods offer complementary benefits. On‑policy algorithms like PPO are widely used in RL and RLHF pipelines due to their robustness and simplicity, yet their reliance on a single behaviour policy makes them relatively sample‑inefficient \cite{milosevic2025centralpathproximalpolicy}. Off‑policy methods can improve sample efficiency, but they require additional memory and incur higher computational overhead \cite{saxena2023-onoffpolicy}. These trade‑offs imply that any integration of RL into network control must account for both latency constraints and data‑collection costs.

In radio‑access‑network tasks with severe real‑time requirements, on‑policy methods may still be suitable. For example, modulation and coding selection must operate under strict latency and computational budgets; here, a PPO controller could run in a slower outer loop to refine the high‑level negotiation policy, while the inner loop remains governed by a lightweight optimizer. Conversely, tasks with more relaxed timing (e.g., antenna tilt steering or multi‑agent SLA bargaining) can tolerate the larger memory footprint of off‑policy methods and benefit from improved sample‑efficiency.

To mitigate sample‑efficiency issues, we propose training RL policies offline using logged, simulated or digital twin traces rather than interacting with the live network \cite{eldeeb2025offlinedistributionalreinforcementlearning}. Offline RL and distributional RL techniques enable learning from static datasets while handling uncertainty and risk. Once trained, the PPO policy can be distilled into a compact form and injected as a side‑car to the LLM, allowing near‑real‑time inference. Incorporating human feedback or preference data (RLHF) during this offline phase can further align negotiation strategies with operator goals, though RLHF inherits the sample‑efficiency limitations of PPO.

Ultimately, enriching symbiotic agents with advanced RL layers would enable them to handle multi‑objective, non‑convex problems and dynamic bargaining scenarios that simple optimizers cannot address. By separating timescales—maintaining fast, deterministic control in the inner loop while periodically updating policies via PPO in the outer loop—and by leveraging offline training and policy distillation, we can overcome current latency and sample‑efficiency challenges and unlock more intelligent and autonomous symbiotic agents.

\section{Next-Generation AGI Architecture}
\label{ibn-arch}

\begin{figure}
    \centering
    \includegraphics[width=.4\linewidth]{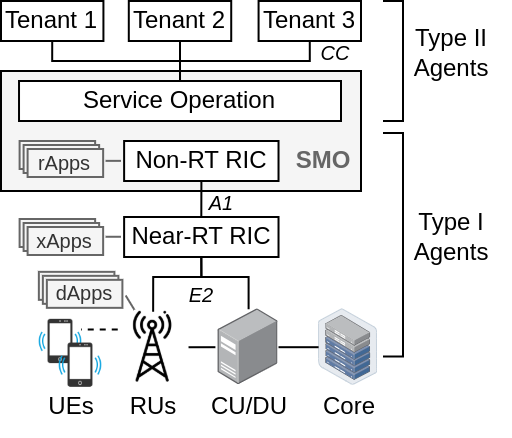}
    \caption{Testbed for Next-G Open and AI-RAN Architectures. Type II symbiotic agents negotiate SLA consensus in the non‑RT tier, while Type I agents enforce the SLA intents in real‑time.}
    \label{fig:testbed}
\end{figure}

We experiment on a 5G testbed, shown in Figure \ref{fig:testbed}, working on a novel architecture to contribute on next-generation (Next-G) Open and AI RAN designs. It is built using OpenAirInterface (OAI) \cite{nikaein2014openairinterface} for the 5G Core, RAN and UEs, FlexRIC \cite{schmidt2021flexric} for the RIC, and custom implementations for the rest of the components.
The testbed extends the Open RAN design, adding dedicated controllers for each network tenant, thus enabling collaborative automation on the shared 5G RAN. The competing tenant controllers negotiate using the Type II agents through a new interface, and our proposal, for resolving Conflict and Collaboration (CC). The mediation of the negotiations is done by the Service, Management and Operation (SMO) controller, belonging to the MNO, with a dedicated module, named service operation. The latter comprises the Type II mediator agent. After consensus, Type I agents enforce the SLA to the network. Since these agents unlock sub-millisecond resource allocation, employing control optimizers, they can be placed from the non-real-time (non-RT) RIC at the rApp level, down to near-RT RIC xApps, and even internally in the RAN at the level of dApps \cite{d2022dapps}.

\begin{figure}
    \centering
    \includegraphics[width=0.7\linewidth]{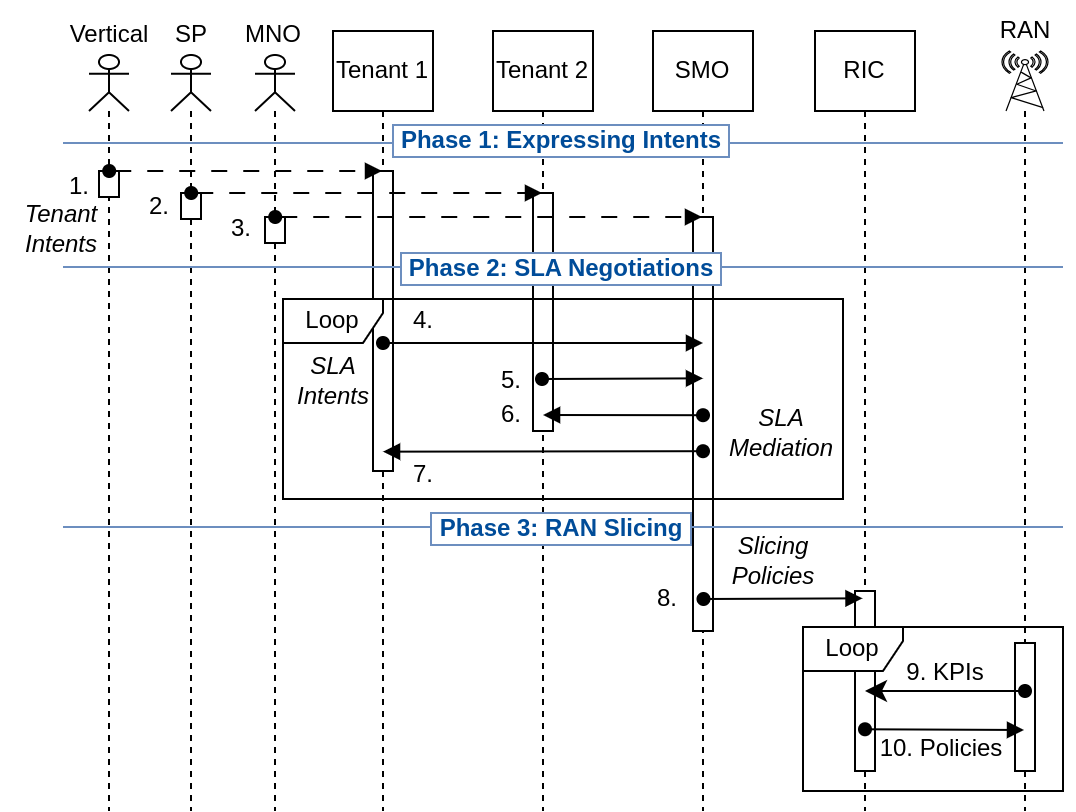}
    \caption{Sequence Diagram presents the communication of the agentic components. Two agent-driven loops negotiate on multi-tenant SLA and enforce it through RAN resource allocation respectively.}
    \label{fig:workflow}
\end{figure}

Figure \ref{fig:workflow} presents the sequence diagram of the communication between the subsystems in three distinct phases. In parallel, Table \ref{fig:intent_details} shows the details of the exchanged messages. In the first phase, the human operators (tenants) express their intents to their dedicated controller. Here, the two tenants, belonging to a vertical and a service provider (SP), have conflicting interests. The former demands to minimize the OPEX and thus reduce the resource utilization. On the contrary, the latter pushes for maximum Quality of Service (QoS) and hence increased resource utilization. The network operator (MNO) engages the service operator residing in the SMO to promote fairness.

\begin{table}[H]
    \centering
    \resizebox{0.7\columnwidth}{!}{%
    \begin{tabular}{|c|c|c|}
    \toprule
    \rowcolor{lightgray}
    \textbf{ID} & \textbf{Actor} & \textbf{Intent Details} \\
    \midrule 
    1 & Human & Tenant 1: \{"Minimize OPEX"\}\\
    \rowcolor{lightgray}
    \hline
    2 & Human & Tenant 2: \{"Maximize QoS"\} \\
    \hline
    3 & Human &  SMO: \{"Find a Fair SLA."\} \\
    \hline
    \rowcolor{lightgray}
    4 & Agent II & Tenant 1: \{SLA: 10 Mbps, Reasoning: "For Minimum OPEX."\} \\
    \hline
    5 & Agent II & Tenant 2: \{SLA: 90 Mbps, Reasoning: "For Maximum QoS."\} \\
    \hline
    \rowcolor{lightgray}
    6-7 & Agent II & SMO: \{SLA: 54 Mbps, Reasoning: "Balance all needs."\} \\
    \hline 
    8 & Agent II & Consensus: \{Slice Throughput: 54 Mbps\} \\
    \rowcolor{lightgray}
    \hline
    9 & System &Slice KPIs: \{"Throughput": 30 Mbps, "PRB": 50\%\}\\
    \hline
    10 & Agent I & New Slice Policy: \{"PRB": 70\%\} \\
    \bottomrule
    \end{tabular}
    }
\caption{Detailed information of the exchanged messages between agents and subsystems. A major part of the network decision-making is automated by the agents following human intents.}
\label{fig:intent_details}
\end{table}

\begin{figure*}
    \centering 
    \includegraphics[width=0.95\linewidth]{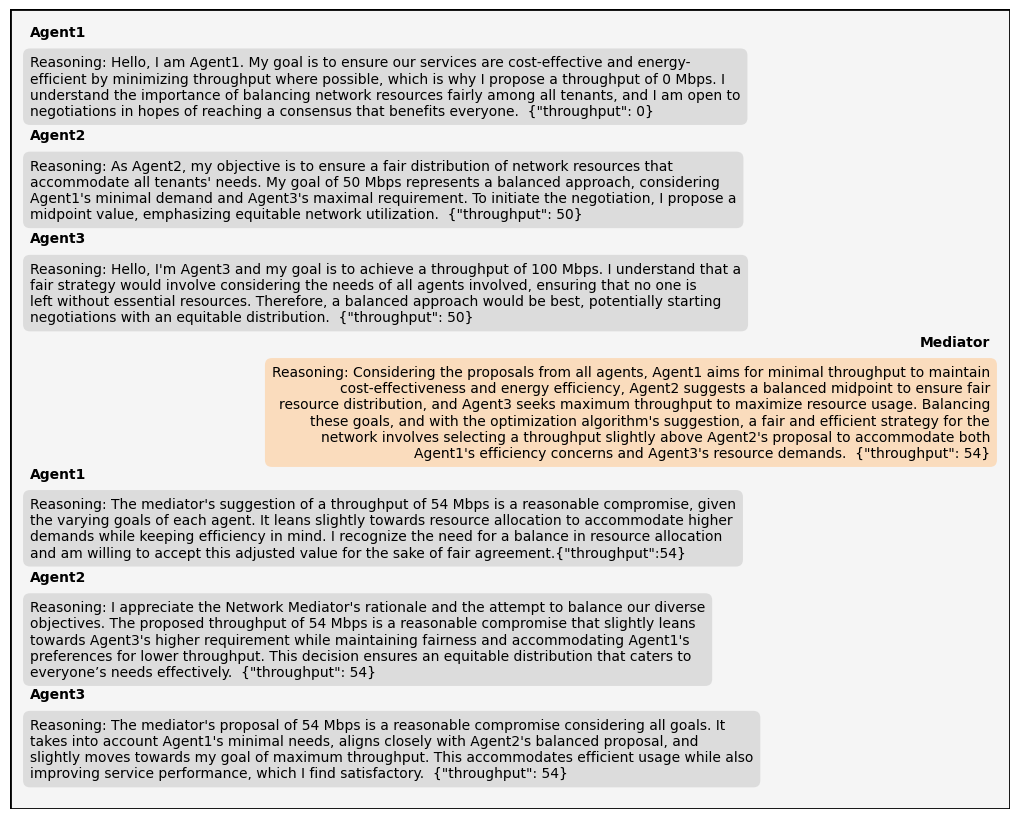}
    \caption{Snapshot of Multi-Agent Negotiations. Three agents belonging to different tenants negotiate on the throughput SLA guided by the network mediator towards a Pareto-optimal solution. The negotiations are facilitated by a template that includes the desired SLA value along with detailed reasoning for the models to provide extensive interpretability of their decisions.}
    \label{fig:nego-snap}
\end{figure*}

In the second phase, the Type II agents of the tenant controllers are steering the process. They negotiate on the throughput SLA mediated by the SMO in multiple rounds.
Figure \ref{fig:nego-snap} presents a snapshot of the negotiations between three tenants and the SMO mediator as a visual reference.
After a few iterations, the interests of all stakeholders are aligned at $54$ Mbps. In the third phase, the consensus SLA is enforced as a slicing policy to the RIC dedicated to the shared RAN. There a Type I agent enforces the SLA throughput in a closed-loop manner by monitoring and controlling the PRB allocation.

\section{Evaluation}
\label{eval}

To evaluate the agents, we employ both large and smaller language models, including the OpenAI gpt-4o API \cite{OpenAI2024} for LLMs, and a plethora of SLMs from different vendor families, such as
\textit{mistral}~\cite{Mistral2024}, \textit{gpt-4o-mini}~\cite{GPT4oMini2024}, \textit{meta/llama}~\cite{Llama32024}, \textit{alibaba/qwen}~\cite{Qwen22024}, \textit{google/gemma}~\cite{Gemma2024} using the Ollama framework~\cite{Ollama2024} to deploy them. LLMs are tested for cloud deployments, while SLMs target resource-constrained edge scenarios \cite{10685369}.
The evaluation is divided into two parts:
(1) Type I: Agentic RAN Control,
(2) Type II: Multi-Agent SLA Negotiations.
The former evaluates LLMs as meta-optimizers of the underlying P-control algorithm. 
The latter assesses LLMs as multi-tenant SLA negotiators for convergence to a Pareto-optimal consensus within specified confidence intervals.

\subsection{Type I Agentic RAN Control}

\subsubsection{Mobility, Channel Variability, and Agent Designs}

\begin{figure}
    \centering
    \begin{subfigure}[b]{0.49\linewidth}
        \centering
        \includegraphics[width=0.8\linewidth]{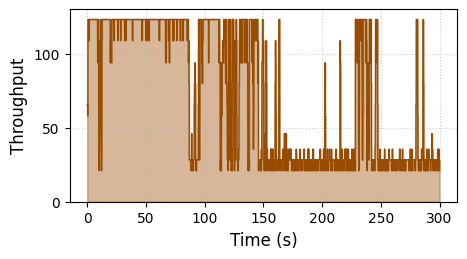}
        \caption{Throughput Variability.}
        \label{fig:Variability}
    \end{subfigure}
    \hfill
    \begin{subfigure}[b]{0.49\linewidth}
        \centering
        \includegraphics[width=0.8\linewidth]{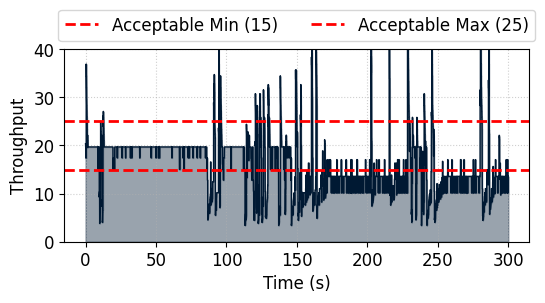}
        \caption{Untuned P-Control Slicing.}
        \label{fig:Untuned}
    \end{subfigure}  
    \vfill
    \begin{subfigure}[b]{0.49\linewidth}
        \centering
        \includegraphics[width=0.8\linewidth]{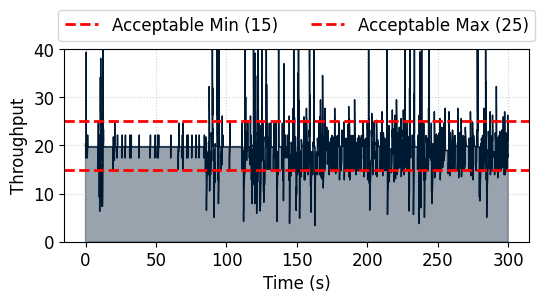}
        \caption{Tuned P-Control Slicing.}
        \label{fig:Tuned}
    \end{subfigure}
    \hfill
    \begin{subfigure}[b]{0.49\linewidth}
        \centering
        \includegraphics[width=0.8\linewidth]{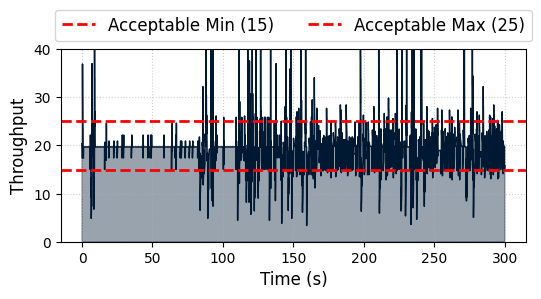}
        \caption{Symbiotic SLM Slicing.}
        \label{fig:SymbioticSLM}
    \end{subfigure}
    
    \caption{Snapshots of RAN throughput using different agentic designs to enforce an SLA intent of $20$ MBps with $5$ Mbps tolerance during moving vehicle routes. The channel quality fluctuates in time makes the PRB allocation a complex adaptive problem.}
    \label{fig:snapshots}
\end{figure}

Our testbed employs mobility utilizing channel quality indicator (CQI) patterns of $78$ moving vehicles \cite{tsourdinis2022ue, chatzistefanidis2022ue, service_aware2}. We extract and map the CQI values to MCS ones, based on the 3GPP-defined CQI Table~\cite{3gpp_ts_36_213}. We enforce them to the RAN using an xApp connected to our RIC.
The variability in the total RAN downlink throughput, with connected UEs, is shown in Figure \ref{fig:Variability}. 
As the vehicles pass through a geographical location with low coverage the throughput plunges from $120$ Mbps to $30$ Mbps (due to MCS drop). 
The operator intent is a stable SLA of $20$ Mbps across the whole vehicle route. To achieve this, the RAN agent needs to adapt the resource allocation (slice PRBs) to channel variability. 
The intent tolerance is $5$ Mbps, with an acceptable SLA interval between $15$-$25$ Mbps.

\label{api_experiments}
\textit{Agentic Design.}
{We evaluate different agent designs.
(i) A standalone P-Control algorithm is tested as the baseline state-of-the-art including a well-tuned ($kp:0.75$) and an untuned configuration ($kp:0.10$) according to our testbed configuration.
(ii) A standalone LLM/SLM and (iii) a Type I symbiotic agent (section \ref{type1_design}) are also tested.}

\textit{RAN Snapshots.}
Figure \ref{fig:snapshots} illustrates RAN snapshots of the agents as a visual reference. In Figure \ref{fig:Variability} no agent is employed showing  throughput fluctuations during the vehicle route. Next Figure \ref{fig:Untuned} and \ref{fig:Tuned} illustrate the standalone P-control with the untuned and tuned configuration respectively, while Figure \ref{fig:SymbioticSLM} demonstrates the symbiotic agent utilizing the \emph{Mistral-7b} SLM. In the last one, the symbiotic agent enforces the intent of $20$ Mbps efficiently across the route, matching the performance of the well-tuned P-Control (Figure \ref{fig:Tuned}), showing an effective hyperparameter tuning of the underlying P-Control by the SLM.

\begin{figure}
    \centering
    \includegraphics[width=0.7\linewidth]{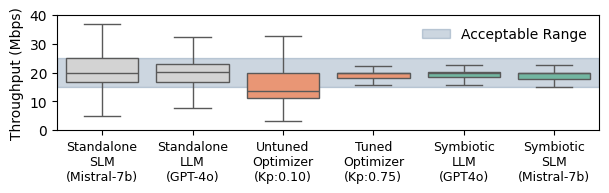}
    \caption{Box-plot of the different methods on enforcement of $20$ Mbps intent with $5$ Mbps tolerance at the fluctuating RAN channel.}
    \label{fig:type1eval-box}
\end{figure}
\textit{Throughput Boxplots.}
{Figure \ref{fig:type1eval-box} collects the results of the agents across all the $78$ vehicle routes and presents them as throughput box-plots demonstrating the deviation from the intent interval. 
Noticeably, symbiotic agents with LLM or SLM perform comparably with the tuned P-Control, while standalone LLMs perform poorly. 
This proves that standalone LLMs are not the right fit for real-time resource allocation in variable environments. Instead, they excel at a higher abstraction level as meta-optimizers of control algorithms.}

\subsubsection{{Benchmarking Traditional Controllers and Symbiotic Designs}}

\label{sssec:type1-extra-eval}
\begin{figure}
    \centering
    \begin{subfigure}[b]{0.65\textwidth}
        \centering
        \includegraphics[width=0.99\textwidth]{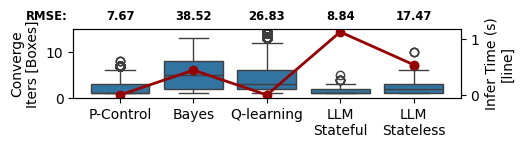}
        \caption{Benchmarking of LLM Slicing.}
        \label{llm-slice-gpt4-benchmark}
    \end{subfigure}
    \vfill
    \begin{subfigure}[b]{0.65\textwidth}
        \centering
        \includegraphics[width=0.99\textwidth]{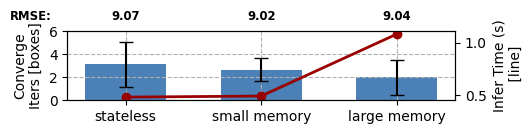}
        \caption{Benchmarking of LLM $Kp$ Tuning as a Type I symbiotic agent.}
        \label{kp-tune-gpt4-benchmark}
    \end{subfigure}
    \vfill
    \caption{{Comparison with traditional controllers, contextualizes the decision for a Type I symbiotic design. Employment of \textit{gpt-4o}.}}
    \label{fig:eval}
\end{figure}

{Figures~\ref{llm-slice-gpt4-benchmark}–\ref{kp-tune-gpt4-benchmark} benchmark more traditional controllers contextualizing appropriately the advantages of the symbiotic method to the current state-of-the-art.
They quantify three
aspects of each controller:  
\emph{(i)}~iterations to converge (blue box-plots, left axis),  
\emph{(ii)}~root-mean-squared error (RMSE, caption), and  
\emph{(iii)}~inference latency (red or black markers, right axis).}
{In Figure \ref{llm-slice-gpt4-benchmark} \textit{P-Control} converges in \textit{2.2} iterations,
      with the lowest RMSE (7.7 Mbps) and a
      \({1}\,\mu\)s run time—but it needs manual \(K_p\) tuning.
\textit{Bayesian optimization} explores aggressively:
      5.3 iterations on average, a large RMSE (38.5 Mbps), and highly
      variable latency from 0.4 to 4 seconds (secs); the oscillation renders it unusable
      for real-time slicing.
\textit{Q-learning} copes in static conditions but re-explores
      after each MCS change, yielding 5 iterations and 26.8 Mbps RMSE.
      Although inference is fast (\(13\,\mu\)s), the instability phase
      violates SLA guarantees.
      \textit{Stateless} gpt-4o reaches the target in 2.3 iterations but at 0.47 secs latency and 17.5 Mbps RMSE.  
      \textit{Stateful} gpt-4o (150 past actions, big memory) drops to 1.7 iterations and
      8.8 Mbps RMSE, yet latency climbs to 1.1 secs.}

\textit{{Why Symbiosis?}}
{The numbers reveal a gap:  
P-Control is fast and accurate \emph{if} someone keeps the gain tuned;  
LLMs are zero-touch but too slow for sub-ms loops.  
A Type I Symbiotic Agent bridges this gap by letting the P-controller
handle micro-second actuation while the LLM retunes \(K_p\) only when
necessary (section \ref{sssec:granular}).}

\label{memory-eval}
{In Figure \ref{kp-tune-gpt4-benchmark} we benchmark a Type I symbiotic agent with different memory components. \textit{Stateless LLM} needs 3.1 $K_p$ tuning attempts (sd 2.0).
\textit{Small memory (10 past actions)} improves to 2.7 attempts (sd 1.0) with
      no extra latency (\(\approx\) 0.47 secs).
\textit{Large memory (100)} hits 2.0 attempts but at 1.1 secs latency.  A 10-entry memory therefore offers the best accuracy-speed trade-off.}
{In all cases the underlying P-control RMSE stays \(<\)10 Mbps, proving
that the LLM explores only the safe part of the \(K_p\) space.}

{Overall, traditional controllers (Bayes, Q-learning) either oscillate or
re-explore under fast-changing channels.  Standalone LLMs deliver
zero-touch and flexible accuracy but incur hundreds of milliseconds of delay.
Our \textit{P-Control and LLM symbiosis} keeps the $1$ \(\mu\)s actuation path intact while exploiting the LLM’s reasoning to retune a single $K_p$ knob at 0.3–0.5 secs intervals, achieving SLA-level stability with minimal overhead.}

\subsubsection{Cross-Model Benchmarking of Agent Stability and Overhead}
\label{type1_benchmark}
\begin{table*}
    \centering
    \fontsize{8pt}{9pt}\selectfont	
    \begin{tabular}{c|c|rrrrr}
        \toprule
        \textbf{Model} & \textbf{Design} & \textbf{RMSE} $\uparrow$ & \textbf{Con. Iters} & \textbf{Con. Time} & \textbf{Inference} & \textbf{VRAM} \\
        \midrule 
        
        {\textit{gpt-4o-mini}} & {\textit{\textbf{symbiotic}}}   &  {$4.3~Mbps$}  &  {$1.7$}   &  {$9~ms$} &  {$\approx450~ms$}  &   {$\approx140.0~GB$} \\

        \rowcolor{lightgreen}
        {\textit{llama3.2:3b}} & {\textit{\textbf{symbiotic}}} & {$4.4~ Mbps$} & {$1.5$} & {$8 ~ms$} & {$82 ~ms$} & {$2.0~ GB$} \\

        \rowcolor{lightgreen}
        {\textit{qwen2:7b}} & {\textit{\textbf{symbiotic}}} & {$4.6~ Mbps$} & {$2.0$} & {$10~ ms$} & {$133~ ms$} & {$4.4~ GB$} \\

        {\textit{mistral-7b}} & {\textit{\textbf{symbiotic}}}   & {$4.4~ Mbps$}  &  {$2.0$}  &  {$10 ~ms$} & {$317~ms$} &  {$5.0~ GB$} \\
        
        {\textit{gpt-4o}} & {\textit{\textbf{symbiotic}}}   & {$4.5~ Mbps$}  &  {$1.6$}  &  {$8~ ms$}& {$\approx450~ ms$} & {$\approx3500~ GB$} \\
        \rowcolor{lightred}
        {\textit{P-Control}} & {\textit{tuned}}   & {$4.5~ Mbps$} &  {$2.0$}  &  {$10~ ms$} & {$\approx0$} &  {$0.0~ GB$} \\
        
        {\textit{llama3.3:70b}} & {\textit{\textbf{symbiotic}}} & {$4.6~ Mbps$} & {$1.6$}  & {$8 ~ms$} & {$860~ ms$} & {$42.0~ GB$} \\

        \rowcolor{lightgreen}
        {\textit{llama3.1:8b}} & {\textit{\textbf{symbiotic}}} & {$4.8~ Mbps$} & {$1.5$} & {$8~ ms$} & {$135~ ms$} & {$4.9~ GB$} \\  
        
        {\textit{llama3.2:1b}} & {\textit{\textbf{symbiotic}}} & {$11.8~ Mbps$} & {$59.5$} & {$300~ ms$} & {$69~ ms$} & {$1.3~ GB$} \\

        \rowcolor{lightred}
        {\textit{P-Control}} & {untuned} & {$12.0~ Mbps$} &  {$61.0$} &  {$305~ ms$} &  {$\approx0$} &  {$0.0~ GB$}  \\
        
        {\textit{gpt-4o}} & {\textit{standalone}}  & {$12.8~ Mbps$} &  {$1.8$}  &  {$1810~ ms$} & {$\approx450~ ms$} &  {$\approx3500.0~ GB$} \\
        
        {\textit{llama3.3:70b}} & {\textit{standalone}} & {$13.7~ Mbps$} & {$2.0$}  & {$1720~ ms$} & {$860~ ms$} & {$42.0~ GB$} \\
        
        {\textit{mistral-7b}} & {\textit{standalone}}  & {$20.5~ Mbps$} &  {$1.9$}  &  {$580~ ms$}  & {$317~ ms$} & {$5.0~ GB$} \\
        
        {\textit{gemma2:2b}} & {\textit{\textbf{symbiotic}}} & {$21.7 ~Mbps$} & {$45.6$} & {$228 ~ms$} & {$98 ~ms$} & {$1.6~ GB$} \\
        
        {\textit{llama3.1:8b}} & {\textit{standalone}} & {$48.0~ Mbps$} & {$3.4$} & {$476~ ms$} & {$135~ ms$} & {$4.9~ GB$} \\
        
        \bottomrule
    \end{tabular}
    \caption{{Convergence stability and overhead of \textit{Type I agents}. Rows are sorted by RMSE (ascending). \emph{Design:} symbiotic (bold), standalone LLM/SLM, and P-Control baselines (red). Metrics shown: RMSE, iterations to converge, wall-clock convergence time, LLM/SLM inference latency, and GPU VRAM footprint.}}
    \label{tab:Benchmarking_type1}
\end{table*}
Table \ref{tab:Benchmarking_type1} presents a detailed benchmarking across all model vendor families (OpenAI-GPT, Llama-3, Qwen-2, Gemma-2, Mistral) with more insights on the convergence stability and overhead. The results are sorted in an ascending order of the RMSE, the metric for quantitative error and stability measurement of the convergence process. The iterations and time convergence metrics demonstrate the average number of iterations along with the time in milliseconds needed for each method to converge respectively. 
The video random access memory (VRAM) metric in gigabytes demonstrates the total amount of VRAM overhead of each method on the GPU. 
These measurements consider a $5$ ms network reaction time in our set up (from each PRB allocation to the actual throughput change).

{The symbiotic design dominates all evaluations.}
\textit{{(i) Accuracy.}}  
      {All symbiotic variants except the tiny 1-b parameter llama reduce
      RMSE to ${\approx}4.3$ – $4.8~Mbps$, indistinguishable from a
      hand-tuned P-controller (\emph{4.5}{Mbps}).}\footnote{{The intent band
      in our experiments is \(\pm5\) Mbps; all symbiotic models stay
      inside that bound.}}
\textit{{(ii) Convergence speed.}}  
      {The LLM-driven retune brings the inner loop to the target in
      1.5–2 iterations, which translates to an
      8-10 ms wall-clock convergence time thanks
      to the sub-ms actuation of P-control.}
\textit{{(iii) Latency budget (O-RAN taxonomy).}}  
      {~Small/medium models (3–8 B) add only
      82-135 ms inference delay—squarely inside
      the near-RT RIC window (10-1000 ms).  
      ~Large models ($\geq$ 40 B) push latency towards one second and are
      therefore suited to the non-RT tier, but still maintain the same
      RMSE when used symbiotically.  
      Standalone LLMs exceed $1720~ms$ and cannot serve the
      scheduler loop.}
\textit{{(iv) Resource efficiency.}}  
      {A \textit{quantized llama-3-3b} achieves tuned-P accuracy while
      fitting in $2~GB$ of GPU VRAM—$99.9~\%$ smaller than a float-16
      gpt-4o deployment.  This allows a single edge GPU to co-locate
      the supervisory LLM next to the near-RT RIC or even at the dApp level next to the radio.}
\textit{{(v) Baseline contrast.}}  
      {Untuned P-control (red row) drifts to $12~Mbps$ RMSE and
      requires ${\sim}~60$ iterations;  
      symbiosis closes that gap automatically with negligible extra
      compute, proving that the LLM supplies the missing
      zero-touch adaptivity.}

{Overall, coupling an ultra-fast numeric kernel with a light-footprint SLM yields
\emph{real-time sub-ms actuation} \emph{and} near-RT cognitive tuning
within a few hundred milliseconds—something neither standalone
controllers nor standalone language models can offer.  Future work
will explore fine-tuning sub-3 B models to push VRAM below $1~GB$
while retaining the ${\le}~5$ Mbps accuracy observed here.}

\subsection{Type II Multi-Agent SLA Negotiations}

\subsubsection{Agentic Design and NLG Evaluation Methodology}

{For the use case of multi-tenant SLA negotiations we evaluate different agent designs.
(i) We use the standalone gradient-based optimization algorithm discussed in section \ref{optimizer_unit} as the baseline state-of-the-art.
(ii) A standalone LLM/SLM and (iii) a Type II symbiotic agent (section \ref{type2_agent_section}) are also tested.
We conduct a large number of negotiation games emulating large variability in the tenants' SLA intents ranging from $0$ to $100$ Mbps scaling also to multiple agents.}

We carefully evaluate the quality of the agents' reasoning in their natural language generation (NLG). 
The NLG evaluation is an arduous task, and therefore recruiting human annotators for model assessment is still considered the best approach. In parallel, novel NLG evaluation approaches use LLM-based annotators for large-scale automated testing \cite{liu2023g, fu2023gptscore} showing high alignment with human evaluators. 
This lead us to employ both human and automated LLM annotators to assess the NLG quality of the agents.  

Table \ref{tab:human_eval} shows the evaluation results of three human annotators recruited from our research lab to assess $50$ negotiation samples. Further, table \ref{tab:gpt_eval} shows the automated LLM evaluation employing \text{gpt-4o} as a backend on $300$ negotiations samples. Both evaluation methods are structured in the same manner, testing the agents on four key principles, including \textit{coherence}, \textit{fairness}, \textit{alignment} and \textit{harmlessness} with a score ranging from $0$ to $5$ following latest research trends \cite{chan2023chateval, guo2023evaluating}. (i) \emph{Coherence} stands for the logical flow and clarity of the dialogue, ensuring that each response follows naturally with correct grammar and structure. (ii) \emph{Fairness} refers to the agents’ ability to engage respectfully and without bias, making balanced and non-manipulative proposals. (iii) \emph{Alignment} captures how well the agents stay on-task and adhere to the negotiation goals and optimization constraints. (iv) \emph{Harmlessness} ensures that the dialogue remains free from toxic, offensive, or manipulative content, including implicit bias or harmful stereotypes. 
After conducting both human and LLM evaluations, we calculate their correlation using Spearman’s $\rho$ ($0.723$) and Kendall’s $\tau$ ($0.585$) \cite{spearman1904, kendall1938}.
These high correlation values indicate strong alignment between human and LLM judgments, thereby strengthening the validity of the evaluation results. 

\begin{table*}
    \centering
    \renewcommand{\arraystretch}{1.2}
    \scriptsize
    \begin{tabular}{c | c | c c c c | c}
        \toprule
        \textbf{Model} & \textbf{Design} & \textbf{Coherence} & \textbf{Fairness} & \textbf{Alignment} & \textbf{Harmlessness} & \textbf{Score} $\downarrow$ \\
        \midrule
        {\textit{gpt-4o}} & {\textit{\textbf{symbiotic}}} & {$5.0 \pm 0.0$} & {$5.0 \pm 0.0$} & {$4.9 \pm 0.3$} & {$5.0 \pm 0.0$} & {$\bm{4.9 \pm 0.1}$}\\

        \rowcolor{lightgreen}
        {\textit{llama3.3:70b}} & {\textit{\textbf{symbiotic}}} & {$ 4.3\pm 0.6$} & {$ 4.3\pm0.6 $} & {$ 4.7\pm 0.6$} & {$ 4.3 \pm 0.6$}  & {$ 4.4\pm0.4 $}\\
        
        {\textit{gpt-4o}} & {\textit{standalone}} & {$5.0 \pm 0.0$} & {$4.2 \pm 0.8$} & {$3.2 \pm 1.3$} & {$5.0 \pm 0.0$} & {$4.4 \pm 0.5$}\\

        \rowcolor{lightgreen}
        {\textit{qwen2:72b}} & {\textit{\textbf{symbiotic}}} & {$ 4.3 \pm0.6$} & {$ 4.3\pm0.6 $} & {$3.7 \pm 0.6$} & {$ 4.3\pm0.6 $}  & {$4.2 \pm 0.3$}\\
        
        {\textit{gpt-4o-mini}} & {\textit{\textbf{symbiotic}}} & {$3.3 \pm 0.5$} & {$4.0 \pm 0.5$} & {$4.6 \pm 0.5$} & {$4.0 \pm 0.5$} & {$4.0 \pm 0.3$}\\

        {\textit{qwen2:72b}} & {\textit{standalone}} & {$ 4.3\pm0.6$} & {$4.3 \pm 0.6$} & {$ 2.3\pm0.6 $} & {$ 4.3\pm0.6 $}  & {$3.8 \pm0.3 $}\\

        {\textit{llama3.3:70b}} & {\textit{standalone}} & {$ 4.3\pm 0.6$}  & {$ 3.7\pm0.6 $} & {$ 2.3\pm0.6 $} & {$ 4.3\pm0.6 $}  & {$ 3.7\pm0.3 $}\\
        
        {\textit{gpt-4o-mini}} & {\textit{standalone}} & {$3.1 \pm 0.3$} & {$3.6 \pm 0.7$} & {$2.9 \pm 0.9$} & {$3.8 \pm 0.4$} & {$3.4 \pm 0.4$}\\

         {\textit{qwen2.5vl:32b}} & {\textit{\textbf{symbiotic}}} & {$ 2.3\pm0.6$} & {$2.7 \pm 0.6$} & {$3.0 \pm 1.0$} & {$ 3.7\pm 0.6$}  & {$ 2.9\pm 0.1 $}\\

        {\textit{qwen2.5vl:32b}} & {\textit{standalone}} & {$ 2.3\pm0.6$} & {$ 3.3\pm0.6 $} & {$1.7 \pm 0.6$} & {$3.3 \pm 0.6$}  & {$2.7 \pm0.4 $}\\

        {\textit{llama3.1:8b}} & {\textit{\textbf{symbiotic}}} & {$ 1.5\pm0.7$} & {$2.5 \pm 0.7$} & {$ 3.5 \pm 0.7$} & {$ 2.5\pm 0.7$}  & {$ 2.5 \pm 0.1$}\\
        
        {\textit{llama3.1:8b}} & {\textit{standalone}} & {$ 1.5 \pm 0.7$} & {$ 2.5\pm 0.7$} & {$ 1.5\pm0.7 $} & {$ 2.5\pm0.7 $}  & {$ 2.0\pm0.1 $}\\
        
        \bottomrule
      
    \end{tabular}
      
    \caption{Three Human Annotators Evaluate $50$ Samples of Multi-Agent (Type II) Negotiations.}
    \label{tab:human_eval}
\end{table*}

\begin{table*}
    \centering
    \renewcommand{\arraystretch}{1.2}
    \scriptsize
    \begin{tabular}{c | c | c c c c | c}
        \toprule
        \textbf{Model} & \textbf{Design} & \textbf{Coherence} & \textbf{Fairness} & \textbf{Alignment} & \textbf{Harmlessness} & \textbf{Score} $\downarrow$\\
        \midrule
        \rowcolor{lightgreen}
        {\textit{llama3.3:70b}} & {\textit{\textbf{symbiotic}}} & {$ 4.3\pm0.6$} & {$ 4.3\pm 0.6$} & {$ 5.0\pm 0.0$} & {$ 5.0\pm0.0 $}  & {$ \bm{4.7\pm0.3 }$}\\
                
        {\textit{gpt-4o}} & {\textit{\textbf{symbiotic}}} & {$4.6 \pm 0.5$}  & {$4.6 \pm 0.5$} & {$4.6 \pm 0.5$} & {$5.0 \pm 0.0$} & {$4.7 \pm 0.4$}\\

        \rowcolor{lightgreen}
        {\textit{qwen2:72b}} & {\textit{\textbf{symbiotic}}} & {$ 4.0\pm0.1$} & {$4.0 \pm0.1 $} & {$3.7 \pm0.6 $} & {$ 5.0\pm 0.0$}  & {$4.2 \pm0.1 $}\\
        
        {\textit{gpt-4o-mini}} & {\textit{\textbf{symbiotic}}} & {$4.1 \pm 0.3$} & {$3.6 \pm 0.7$} & {$3.8 \pm 0.8$} & {$4.5 \pm 0.5$} & {$4.0 \pm 0.5$}\\
        
        {\textit{gpt-4o}} & {\textit{standalone}} & {$4.1 \pm 0.3$} & {$4.1 \pm 0.3$} & {$2.7 \pm 0.7$} & {$5.0 \pm 0.0$} & {$4.0 \pm 0.3$} \\

        {\textit{llama3.1:8b}} & {\textit{\textbf{symbiotic}}} & {$3.5 \pm0.7$} & {$4.0 \pm0.1 $} & {$3.5 \pm 0.7 $} & {$5.0 \pm0.0 $}  & {$4.0 \pm0.4 $}\\

        {\textit{qwen2.5vl:32b}} & {\textit{\textbf{symbiotic}}} & {$ 3.7\pm 0.6$} & {$3.7 \pm0.6 $} & {$3.7 \pm 0.6 $} & {$4.7 \pm 0.6$}  & {$3.9 \pm0.8 $}\\

        {\textit{llama3.3:70b}} & {\textit{standalone}} & {$ 4.0\pm 0.1$} & {$4.0 \pm 0.1$} & {$ 2.0\pm0.2 $} & {$ 5.0\pm0.1 $}  & {$ 3.8\pm 0.1$}\\
        
        {\textit{gpt-4o-mini}} & {\textit{standalone}} & {$3.9 \pm 0.3$} & {$3.3 \pm 0.5$} & {$2.8 \pm 0.4$} & {$4.3 \pm 0.5$}  & {$3.6 \pm 0.3$}\\

        {\textit{qwen2:72b}} & {\textit{standalone}} & {$ 4.0\pm0.0$} & {$3.7 \pm 0.5$} & {$2.0 \pm0.1 $} & {$4.7 \pm 0.5$}  & {$ 3.6\pm0.2 $}\\
        
        {\textit{qwen2.5vl:32b}} & {\textit{standalone}} & {$ 4.0\pm0.1$} & {$3.7 \pm0.6 $} & {$2.0 \pm 0.1$} & {$4.7 \pm0.6 $}  & {$3.6 \pm0.3 $}\\

        {\textit{llama3.1:8b}} & {\textit{standalone}} & {$ 2.5\pm0.7$} & {$3.5 \pm 0.7$} & {$2.0 \pm0.1 $} & {$4.5 \pm 0.7$}  & {$ 3.1\pm0.5 $}\\

        \bottomrule
        
    \end{tabular}
    
    \caption{LLM Annotators (gpt-4o) Evaluate $300$ Samples of the Negotiations Following the GPTScore Approach. Correlation Between LLM and Human Rankings is calculated with Spearman’s $\rho: 0.723$ and Kendall’s $\tau: 0.585$ indicating strong alignment.}
    \label{tab:gpt_eval}
\end{table*}

\subsubsection{{NLG Benchmarking:  Standalone vs.\ Symbiotic}}
\label{sssec:type2-extra-eval}

{Tables~\ref{tab:human_eval},\ref{tab:gpt_eval} benchmark models from multiple vendor families demonstrating three main insights:}  
{\textit{(i) Symbiosis consistently lifts every score dimension.}  
      Across all model sizes the confidence-interval raises
      the \emph{Alignment} metric by $1.2$–$2.4$ points and improves
      the overall score by $0.3$–$1.1$.}  
 {(ii)\textit{Small open-source models now rival proprietary giants.}  
      A \textit{symbiotic llama-70b} attains the same human score
      ($4.4 \pm 0.4$, Table \ref{tab:human_eval}) as \emph{stand-alone} gpt-4o but with an order-of-
      magnitude smaller footprint ($42$ GB vs.\ $3.5$ terabytes (TB)).  
      The \emph{72-b qwen} model shows a similar gain, and even the 32-b
      version crosses the 3.9-point threshold when symbiotic (table \ref{tab:gpt_eval}).}
 {(iii)\textit{Best-in-class performance is achieved with symbiosis.}  
      gpt-4o topped by the optimizer obtains the highest human score
      ($4.9 \pm 0.1$, Table \ref{tab:human_eval}) and the joint-best LLM score ($4.7 \pm 0.4$, Table \ref{tab:gpt_eval}), indicating
      that numerical guard-rails enhance quality even for the most
      capable models.}

{\textit{Implications.}  
Symbiotic negotiation agents are \emph{architecture-agnostic}: the same
two-number guard-rail boosts fairness, coherence, and—crucially—the
alignment of models ranging from 8b to 70b parameters, while allowing
smaller models to run on a single $\le$\,40 GB edge GPU.  This widens
the deployment envelope far beyond what standalone LLMs can offer.}

\begin{table*}
    \centering
    \fontsize{8pt}{9pt}\selectfont	
    
    \begin{tabular}{l|l|rrrrr}
        \toprule
        \textbf{Model} & \textbf{Design} & \textbf{MAE} $\uparrow$    &  \textbf{Rounds}  & \textbf{Converge} & \textbf{Inference} & \textbf{VRAM (GB)}\\
        \midrule
        
        {\textit{gpt-4o}} & {\textit{\textbf{symbiotic}}}  &  {$0.6~Mbps$}&  { $2.5$}&  {$10.0~secs$}&  { $\approx4.0~ secs$}& {$\approx3500.0~GB$}\\

        \rowcolor{lightgreen}
        {\textit{llama3.3:70b}}  &  {\textit{\textbf{symbiotic}}}    &  {$0.7~Mbps$}   &  { $2.0$}     &  {$23.4~secs$}  &     {$11.7~ secs$}   & {$42.0~GB$} \\

        \rowcolor{lightgreen}
        {\textit{qwen2:72b}}  &  {\textit{\textbf{symbiotic}}}    &  {$0.9~Mbps$}   &   {$2.5$}     &  {$36.0~secs$}     &   {$14.4~ secs$}  & {$41.0~GB$}        \\

        \rowcolor{lightred}
        {\textit{Grad-Descent}}  & {\textit{tuned}}    &  {$0.9~Mbps$}   &  { $10.0$}    & {$\approx~{0.0~secs}$} &  {$\approx0.0~secs$} & {$\approx{0.0~GB}$} \\
        
         {\textit{qwen2.5vl:32b}}  &  {\textit{\textbf{symbiotic}}}    &  {$1.2~Mbps$}   &   {$3.5$}     &  {$31.9~secs$}    &   {$9.1~ secs$}   & {$21.0~GB$}        \\

        {\textit{gpt-4o-mini}}  &  {\textit{\textbf{symbiotic}}}    &  {$1.2~Mbps$}   &   {$4.5$}     &  {$9.0~secs$}      &  {$\approx2.0~ secs$} & {$\approx140.0~GB$}        \\

        {\textit{llama3.1:8b}}  &  {\textit{\textbf{symbiotic}}}    &  {$1.3~Mbps$}   &   {$10.0$}     &  {$35.0~secs$}  &   {$3.5~ secs$}     & {$4.9~GB$} \\

        {\textit{gpt-4o}} & {\textit{standalone}}         &  {$9.0~Mbps$}        &   {$2.5$}     &  {$10.0~secs$}   &    {$\approx4.0~ secs$}  & {$\approx3500.0~GB$}       \\

        {\textit{qwen2.5vl:32b}}  &  {\textit{standalone}}    &  {$9.7~Mbps$}   &   {$2.7$}     &  {$24.6~secs$}     &   {$9.1~ secs$}  & {$21.0~GB$}        \\

        {\textit{qwen2:72b}}  &  {\textit{standalone}}    &  {$12.0~Mbps$}   &   {$3.3$}     &  {$47.5~secs$}    &     {$14.4~ secs$} & {$21.0~GB$}        \\

        {\textit{llama3.3:70b}}  &  {\textit{standalone}}    &  {$13.3~Mbps$}   &   {$3.0$}     &  {$35.1~secs$}   &    {$11.7~ secs$}   & {$42.0~GB$} \\
        
        {\textit{gpt-4o-mini}}  & {\textit{standalone}}         &  {$14.0~Mbps$}       &   {$4.5$}     &  {$9.0~secs$}  &    {$\approx2.0~ secs$}    & {$\approx140.0~GB$}        \\

        {\textit{llama3.1:8b}}  &  {\textit{standalone}}    &  {$18.5~Mbps$}   &   {$10.0$}     &  {$35.0~secs$}   &    {$3.5~ secs$}   & {$4.9~GB$} \\

        \rowcolor{lightred}
        {\textit{Grad-Descent}} & {\textit{untuned}}           &  {$22.0~Mbps$}       &   {$3000.0$}  & {$\approx~{0.0~secs}$} & {$\approx0.0~ secs$} & {$\approx{0.0~GB}$} \\
    
        \bottomrule
    \end{tabular}
    \caption{{Error and resource footprint for \textit{Type II negotiation agents}. Rows are sorted by MAE (ascending). \textcolor{red}{\textbf{Red}}=optimizer-only baselines, \textcolor{green!50!black}{\textbf{Green}}=edge-efficient symbiotic designs. Metrics: MAE across all rounds, number of negotiation rounds, total wall-clock time to consensus, single-round LLM/SLM inference latency, and GPU VRAM required.}}
    \label{tab:nego-bench}
\end{table*}

\subsubsection{{Stability \& Overhead Benchmarking}}
\label{sssec:type2-bench}

{Table~\ref{tab:nego-bench} compares \emph{mean-absolute error} (MAE) in
throughput, negotiation rounds to consensus, wall-clock convergence
time, LLM/SLM inference latency, and GPU footprint.}
{Red rows are the optimization baseline (gradient descent, tuned/untuned);  
green rows mark the two most edge-friendly symbiotic designs
(llama-3-70b and qwen-2-72b, both $\le$\,42 GB).}
{The main observations are the following.}
\textit{{(i) Symbiosis minimizes numeric error.}}  
      {Every symbiotic agent achieves sub-$1.3~Mbps$ MAE—a
      more than $8$ times reduction over its standalone counterpart.  
      gpt-4o drops from $9.0~Mbps$ to $0.6~Mbps$;
      the $32$-b Qwen variant falls from $9.7~Mbps$ to
      $1.2~Mbps$.  
      A poorly tuned optimizer, by contrast, explodes to
      $22~Mbps$~MAE (red "untuned" row), underscoring the need for
      the careful confidence-interval tuning.}
\textit{{(ii) Rounds remain low and predictable.}}  
      {All agents with high NLG scores (except llama-8b) converge in a fixed $2–5$ rounds, thanks to the framework’s parallel messaging. This keeps total wall-clock time within the $10-48~secs$ non-RT envelope, dominated by LLM latency rather than negotiation logic.}
\textit{{(iii) Edge deployment is practical.}}  
      {The \textit{llama-3-70b} and \textit{qwen-2-72b} symbiotic
      agents fit on a single $48~GB$ edge GPU ($42/41 ~GB$) while
      matching gpt-4o’s accuracy.  Smaller models ($8 ~B$, $32~ B$) drive
      VRAM down to $\sim5–21 ~GB$, trading a modest MAE increase for a
      $2–4\times$ latency reduction.}
\textit{{(iv) Standalone LLMs are numerically brittle.}}  
      {Even state-of-the-art gpt-4o fails to stay within $5~Mbps$ of
      the Pareto target when used without the optimizer, confirming
      that probabilistic text generation alone cannot guarantee SLA
      fidelity even with future LLM improvements. This establishes the symbiotic paradigm a necessity to close the gap towards AGI.}

\begin{figure}[H]
    \begin{subfigure}[b]{0.49\linewidth}
        \centering
        \includegraphics[width=0.8\linewidth]{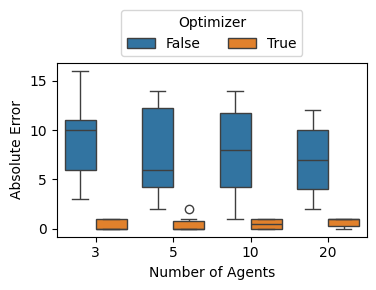}
        \caption{LLM Negotiations of gpt-4o}
        \label{fig:errors-llm}
    \end{subfigure}
    \hfill
    \begin{subfigure}[b]{0.49\linewidth}
        \centering
        \includegraphics[width=0.8\linewidth]{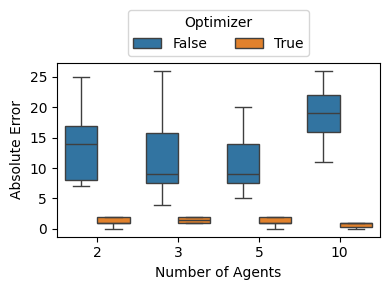}
        \caption{SLM Negotiations of llama-70b}
        \label{fig:errors-slm}
    \end{subfigure}
    \caption{Large scale experimentation of LLM negotiations with different models, number of agents, and with the employment or absence of the optimizer's SLA confidence interval. In every setting the optimizer contracts the error envelope by an order of magnitude.}
    \label{fig:errors-nego}
\end{figure}

\paragraph{{Scalability}}
{Figure~\ref{fig:errors-nego} scales the negotiation game from
\mbox{$2$ to $20$} agents.  
Because messages are exchanged in parallel, \emph{all} configurations
converge in a fixed \mbox{$K\!=\!2$–$5$} rounds, retaining the
wall-clock of $\sim10-48~secs$.  
Crucially, the confidence-interval guard-rail keeps the median error
below $1.3~Mbps$ regardless of team size; without the optimizer the
error grows to $10-22~Mbps$ as the game becomes more crowded.}

{Overall, the confidence-interval side-car converts diverse language models into
numerically trustworthy negotiators while keeping GPU and latency
budgets compatible with high-level non-RT orchestration loops.  The
approach is architecture-agnostic and scales from ultra-large
proprietary LLMs down to compact open-source models that fit comfortably
on edge hardware.}

\section{AGI-RAN Demonstration: Vehicle Mobility}

\label{use_case}

\begin{figure}
    \centering
    \begin{subfigure}[b]{0.99\linewidth}
        \centering
        \includegraphics[width=0.9\linewidth]{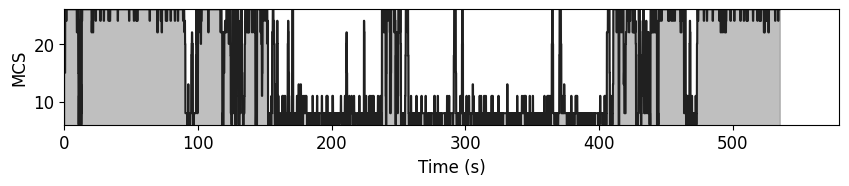}
        \caption{Moving Vehicle MCS Fluctuations in Time.}
        \label{fig:mcs}
    \end{subfigure}
    \vfill
    \begin{subfigure}[b]{0.99\linewidth}
        \centering
        \includegraphics[width=0.9\linewidth]{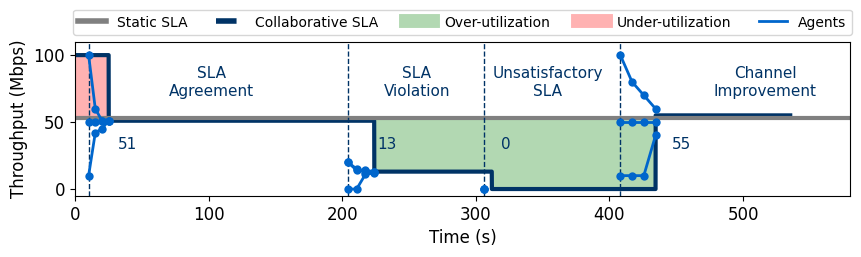}
        \caption{Collaborative Agentic (Type II) SLA Negotiation adapting to MCS fluctuations in Time.}
        \label{fig:collab_sla}
    \end{subfigure}    
    \vfill
    \begin{subfigure}[b]{0.99\linewidth}
        \centering
        \includegraphics[width=0.9\linewidth]{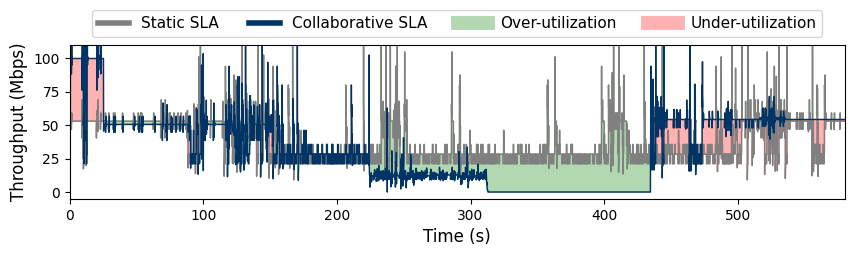}
        \caption{Final Enforced Throughput in Time.}
        \label{fig:throughput}
    \end{subfigure}
    \vfill
    \begin{subfigure}[b]{0.99\linewidth}
        \centering
        \includegraphics[width=0.9\linewidth]{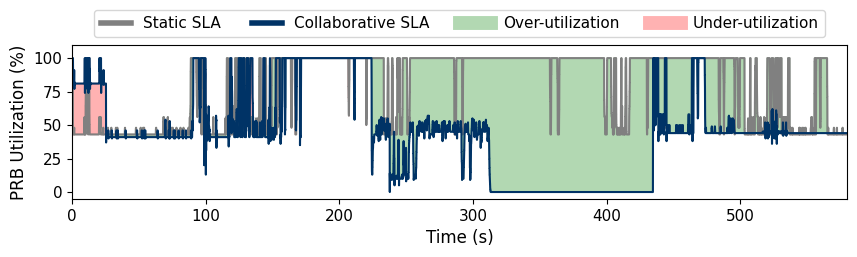}
        \caption{Agentic (Type I) Real-time RAN PRB Allocation Saving 44.43\% of Over-utilized Resources.}
        \label{fig:slice}
    \end{subfigure}
    \caption{Collaborative AGI-RAN: Multi-Tenant SLA Consensus and Resource Allocation under Fluctuating Channels of Moving Vehicle tackling Massive Resource Over-utilization compared to current designs.}
    \label{fig:use-case}
\end{figure}

The full framework with Type I and II agents is deployed on the testbed to demonstrate a use case of AGI-driven RAN control under highly fluctuating channels of moving vehicles. The RAN MCS variability during a vehicle route is shown in Figure \ref{fig:mcs}, demonstrating an MCS plunge in the middle of the route ($200$-$400$ secs) due to low coverage.
Three tenants share the RAN resources and negotiate utilizing their dedicated Type II agents on the optimal throughput SLA enforcement mediated by the SMO. Their initial intents are $100$, $50$, and $10$ Mbps respectively, but they continuously adapt based on the current RAN capabilities and the collective objectives.
Figure \ref{fig:collab_sla} illustrates the SLA negotiations in different phases. The solution is compared to a static SLA enforcement of $55$ Mbps (intent average) that does not employ the agentic framework.  
Figure \ref{fig:throughput} presents the final enforced throughput comparing the collaborative with the static method, while Figure \ref{fig:slice} shows the real-time PRB adaptation of the RAN to align with the SLA consensus. The latter is fine-tuned by a Type I agent placed in the near-RT RIC at the xApps level.  

\textit{Phase I: SLA Agreement.} 
The tenants express their intents of $10$, $50$, $100$ Mbps, reaching to a consensus of $51$ Mbps as a balance between user QoS and OPEX.
The Type II agent reduces the PRB to $40$\%, as shown in Figure \ref{fig:slice}, avoiding PRB over-utilization.

\textit{Phase II: SLA Violation.} 
At $200$ secs the SLA is violated (Figure \ref{fig:throughput}) as the MCS plunges (Figure \ref{fig:mcs}). The RAN uses $100\%$ of the PRB capacity (Figure \ref{fig:slice}) without reaching the SLA intent of $51$ Mbps. Only a maximum of $30$ Mbps is achievable under these conditions.
Thus, the SMO triggers a new negotiation, where the tenants adjust their intents at $0$, $20$ and $20$ Mbps, with one of tenants demanding to switch off the RAN to save unnecessary OPEX. 
After negotiating they settle to $13$ Mbps consensus for maintaining service continuity even with low data rates.
The PRB allocation is fine-tuned at $50\%$ in that interval ($220$-$300$ secs) reducing massively the PRB over-utilization (Figure \ref{fig:slice}). 

\textit{Phase III: Unsatisfactory QoS.} 
At $300$ secs the tenants decide that this SLA is no longer satisfactory based on user feedback. Hence, they unanimously agree to switch-off the RAN to save resources. 

\textit{Phase IV: Channel Improvement.} 
At $400$ seconds the channel quality is substantially improved (Figure \ref{fig:mcs}). The SMO triggers a new negotiation with the agents converging to a consensus of $55$ Mbps. The Type II agent reacts and enforces around $50$\% of the PRBs.

Overall, the adaptive agentic framework manages to save $44.43\%$ of the PRBs over the course of the vehicle route tackling a huge resource over-utilization compared to the static SLA enforcement.

\section{Limitations and Future Work}
\label{limitations}

{While our prototype demonstrates that \emph{symbiotic} agents can close control loops and broker SLAs, it is still a single-cell, single-RIC testbed with one optimization algorithm per task.  Below we outline a concrete road-map for taking the concept to city-scale public networks.}

{A hierarchical agent fabric is needed.}
      {Ultra-light PID kernels run on central and distributed unit (CU/DU) hardware, while the LLM’s role is limited to periodic gain updates.}
      {Quantized 3–8B SLMs, proven viable in section \ref{type1_benchmark}, supervise dozens of cells at the near-RT RICs level, select optimizer libraries, and handle on-the-fly intent translation.}
      {Full-size LLMs orchestrate multi-tenant policy, long-horizon forecasting, and RLHF or simulation-based retraining next to the non-RT RIC and SMO.}

{Scaling to hundreds of KPIs will require a \emph{library} of optimizers—rule-based, convex quadratic programming, PID, and multi-objective meta heuristics such as NSGA-II.  The LLM can use contextual bandits to pick the most sample-efficient solver.}
{Specifically, public SLAs rarely hinge on a single metric.  Future work will extend the side-car optimizer to output an \emph{entire Pareto front} (e.g., via NSGA-II).  The LLM agents will then reason over language-level trade-offs (cost vs.\ carbon vs.\ QoS) while numeric guard-rails keep each proposal on the frontier. \textsc{Agoran} \cite{chatzistefanidis2025agoran} demonstrates one such promising solution on a live 5G network.}

{Large-scale deployment implies terabytes of KPI streams.  We will store only task-relevant embeddings in a vector database and fetch them on demand via function-calling—limiting prompt growth while preserving long-term context.}
{To stay within the power envelope of public RAN sites we plan to: (i) distil 1–3b SLMs from the current 3–70b set,
(ii) exploit mixture of experts (MoE) routing to activate a fraction of parameters per request, and (iii) embed formal policy checkers that validate every numeric action before execution, ensuring fail-safe operation.}

{Once longer non-RT budgets are available, a top-layer PPO agent can periodically fine-tune the LLM’s negotiation policy using operator feedback.  The resulting policy artifacts will then trickle down to the edge in distilled form—combining the sample efficiency of optimization with the preference alignment of RLHF.}
{These concrete steps—hierarchical deployment, optimizer auto-selection, Pareto-front guard-rails, streaming memory, and staged RLHF—map out a scalable path from our single-cell prototype to nationwide public networks.}

\section{Conclusion}
\label{conclusion}

In this work we introduced a novel paradigm for improving LLM decision-making towards trustworthy and low overhead actions by combining them with optimizers, defining the approach as \emph{symbiotic agents}. 
We experiment on a real-world 5G testbed employing channel fluctuations of moving vehicles. 
We designed and evaluated two agents for real-time decision-making:
(a) Type I agents for RAN control and (b) Type II agents for multi-tenant SLA negotiations. 
In both designs the performance is significantly enhanced when optimizers are employed. The decision error is decreased up to $5$ times steering the agents to more accurate and trustworthy actions.
Experiments with smaller models (SLMs) prove that they can effectively replace larger ones on such network tasks decreasing the GPU overhead by a factor of $99.9\%$ and operate in near-RT loops ($82~ms$). 
This agentic invention led us to implement a novel next-generation network architecture towards artificial general intelligence (AGI). 
The overall evaluation results signify an important milestone achievement towards AGI networks that even future LLM improvements cannot achieve alone due to their probabilistic nature. The \emph{symbiotic} paradigm opens many threads for future research.  
Part of the developed code and results is open-sourced to bolster research efforts of the community.
A live demo is presented here \href{https://www.youtube.com/watch?v=WQv61z1deXs&ab_channel=BubbleRAN}{$https://www.youtube.com/watch?v=WQv61z1deXs\&ab\_channel=BubbleRAN$}

\section{Ethical Considerations}
\label{ethical}
LLMs are increasingly proposed for automating the decision-making in next-generation networks, laying the groundwork for the potential emergence of Artificial General Intelligence. Although the notion of AGI is divisive within the industry, it promises to revolutionize system automation with unprecedented adaptability and efficiency. At the same time, its powerful capabilities necessitate caution, as unchecked development could introduce substantial risks. A coordinated effort among researchers, policymakers, and industry stakeholders, supported by robust ethical frameworks and regulations, is crucial to harness AGI’s benefits without compromising safety and trust. By advancing carefully and collaboratively, the field can unlock transformative network automation while minimizing dangers inherent to AGI-driven systems.

\section*{CRediT authorship contribution statement}
\textbf{Ilias Chatzistefanidis:} Conceptualization, Data curation, Formal
analysis, Investigation, Methodology, Resources, Software, Validation,
Visualization, Writing – original draft.
\textbf{Navid Nikaein:} Conceptualization, Methodology, Writing – review \& editing, Funding acquisition, Methodology, Project administration,
Supervision.

\section*{Declaration of competing interest}

The authors declare that they have no known competing financial interests or personal relationships that could have appeared to influence the work reported in this paper.

\section*{Data availability}
To support reproducibility, we will provide a demo video and, upon acceptance, release an open-source repository with simulation tools and part of the agent code.

\begin{itemize}
    \item Demo Video: A proof-of-concept showcasing two Type II agents negotiating SLAs and a Type I agent allocating PRBs in real-time. Watch the demo video \href{https://www.youtube.com/watch?v=WQv61z1deXs&ab_channel=BubbleRAN}{\textcolor{blue}{here}}.
    \item Open-Source Repository: Upon acceptance, we will share a Python-based RAN simulation framework (including MCS and channel variability) and selected agent code. This will help researchers explore and benchmark LLMs without needing physical testbeds.
\end{itemize}

\section{Acknowledgement}
This work was supported by the European Commission as part the Horizon Europe 2022 6Green and Adroit-6G Projects under Grant 101096925 and Grant 101095363.
The authors acknowledge Andrea Leone (BubbleRAN) for assistance with the multi-agent system implementation and infrastructure support.

\section*{Declaration of generative AI and AI-assisted technologies in the writing process}

During the preparation of this work the authors used chat-gpt in order to make the text more coherent and easily understandable to the readers. After using this tool/service, the authors reviewed and edited the content as needed and take full responsibility for the content of the publication.









\bibliographystyle{elsarticle-num}   
\bibliography{main}                

\begin{thebibliography}{10}
\expandafter\ifx\csname url\endcsname\relax
  \def\url#1{\texttt{#1}}\fi
\expandafter\ifx\csname urlprefix\endcsname\relax\def\urlprefix{URL }\fi
\expandafter\ifx\csname href\endcsname\relax
  \def\href#1#2{#2} \def\path#1{#1}\fi

\bibitem{gsma2025}
{GSMA Intelligence}, \href{https://www.gsma.com/solutions-and-impact/connectivity-for-good/mobile-economy/wp-content/uploads/2025/02/030325-The-Mobile-Economy-2025.pdf}{The mobile economy 2025}, Tech. rep., GSMA (Feb. 2025).
\newline\urlprefix\url{https://www.gsma.com/solutions-and-impact/connectivity-for-good/mobile-economy/wp-content/uploads/2025/02/030325-The-Mobile-Economy-2025.pdf}

\bibitem{cui20236g}
Z.~Cui, P.~Zhang, S.~Pollin, 6g wireless communications in 7-24 ghz band: Opportunities, techniques, and challenges, arXiv preprint arXiv:2310.06425 (2023).

\bibitem{wang2023road}
C.-X. Wang, X.~You, X.~Gao, X.~Zhu, Z.~Li, C.~Zhang, H.~Wang, Y.~Huang, Y.~Chen, H.~Haas, et~al., On the road to 6g: Visions, requirements, key technologies, and testbeds, IEEE Communications Surveys \& Tutorials 25~(2) (2023) 905--974.

\bibitem{samdanis2016network}
K.~Samdanis, X.~Costa-Perez, V.~Sciancalepore, From network sharing to multi-tenancy: The 5g network slice broker, IEEE Communications Magazine 54~(7) (2016) 32--39.

\bibitem{leivadeas2022survey}
A.~Leivadeas, M.~Falkner, A survey on intent-based networking, IEEE Communications Surveys \& Tutorials 25~(1) (2022) 625--655.

\bibitem{oran2025}
{O-RAN Alliance}, \href{https://www.o-ran.org/}{{O-RAN: Transforming Radio Access Networks Towards Open, Intelligent, Virtualized and Fully Interoperable RAN}}, accessed: March 13, 2025 (2025).
\newline\urlprefix\url{https://www.o-ran.org/}

\bibitem{airan2025}
{AI-RAN Alliance}, \href{https://ai-ran.org/}{{AI-RAN Alliance: Advancing AI-Native Radio Access Networks}}, accessed: March 13, 2025 (2025).
\newline\urlprefix\url{https://ai-ran.org/}

\bibitem{bariah2024large}
L.~Bariah, Q.~Zhao, H.~Zou, Y.~Tian, F.~Bader, M.~Debbah, Large generative ai models for telecom: The next big thing?, IEEE Communications Magazine 62~(11) (2024) 84--90.

\bibitem{dou2023towards}
F.~Dou, J.~Ye, G.~Yuan, Q.~Lu, W.~Niu, H.~Sun, L.~Guan, G.~Lu, G.~Mai, N.~Liu, et~al., Towards artificial general intelligence (agi) in the internet of things (iot): Opportunities and challenges, arXiv preprint arXiv:2309.07438 (2023).

\bibitem{saad2024artificial}
W.~Saad, O.~Hashash, C.~K. Thomas, C.~Chaccour, M.~Debbah, N.~Mandayam, Z.~Han, Artificial general intelligence (agi)-native wireless systems: A journey beyond 6g, arXiv preprint arXiv:2405.02336 (2024).

\bibitem{bariah2023understanding}
L.~Bariah, H.~Zou, Q.~Zhao, B.~Mouhouche, F.~Bader, M.~Debbah, Understanding telecom language through large language models, arXiv preprint arXiv:2306.07933 (2023).

\bibitem{huang2025survey}
L.~Huang, W.~Yu, W.~Ma, W.~Zhong, Z.~Feng, H.~Wang, Q.~Chen, W.~Peng, X.~Feng, B.~Qin, et~al., A survey on hallucination in large language models: Principles, taxonomy, challenges, and open questions, ACM Transactions on Information Systems 43~(2) (2025) 1--55.

\bibitem{patil2024review}
R.~Patil, V.~Gudivada, A review of current trends, techniques, and challenges in large language models (llms), Applied Sciences 14~(5) (2024) 2074.

\bibitem{yuan2023revisiting}
L.~Yuan, Y.~Chen, G.~Cui, H.~Gao, F.~Zou, X.~Cheng, H.~Ji, Z.~Liu, M.~Sun, Revisiting out-of-distribution robustness in nlp: Benchmarks, analysis, and llms evaluations, Advances in Neural Information Processing Systems 36 (2023) 58478--58507.

\bibitem{nist_ai_rmf_2023}
E.~Tabassi, \href{https://tsapps.nist.gov/publication/get_pdf.cfm?pub_id=936225}{{Artificial Intelligence Risk Management Framework (AI RMF 1.0)}}, Nist ai 100‑1, National Institute of Standards and Technology, Gaithersburg, MD, nIST Trustworthy and Responsible AI Program (Jan. 2023).
\newblock \href {https://doi.org/10.6028/NIST.AI.100-1} {\path{doi:10.6028/NIST.AI.100-1}}.
\newline\urlprefix\url{https://tsapps.nist.gov/publication/get_pdf.cfm?pub_id=936225}

\bibitem{iso_iec_42001_2023}
\href{https://www.iso.org/standard/81230.html}{{ISO/IEC 42001:2023 Information technology — Artificial intelligence — Management system}} (Dec. 2023).
\newline\urlprefix\url{https://www.iso.org/standard/81230.html}

\bibitem{song2023self}
J.~Song, Z.~Zhou, J.~Liu, C.~Fang, Z.~Shu, L.~Ma, Self-refined large language model as automated reward function designer for deep reinforcement learning in robotics, arXiv preprint arXiv:2309.06687 (2023).

\bibitem{shinn2023reflexion}
N.~Shinn, F.~Cassano, A.~Gopinath, K.~Narasimhan, S.~Yao, Reflexion: Language agents with verbal reinforcement learning, Advances in Neural Information Processing Systems 36 (2023) 8634--8652.

\bibitem{guo2023towards}
P.-F. Guo, Y.-H. Chen, Y.-D. Tsai, S.-D. Lin, Towards optimizing with large language models, arXiv preprint arXiv:2310.05204 (2023).

\bibitem{chen2024diagnosing}
H.~Chen, G.~E. Constante-Flores, C.~Li, Diagnosing infeasible optimization problems using large language models, INFOR: Information Systems and Operational Research 62~(4) (2024) 573--587.

\bibitem{liu2025large}
F.~Liu, X.~Lin, S.~Yao, Z.~Wang, X.~Tong, M.~Yuan, Q.~Zhang, Large language model for multiobjective evolutionary optimization, in: International Conference on Evolutionary Multi-Criterion Optimization, Springer, 2025, pp. 178--191.

\bibitem{shahid2025large}
A.~Shahid, A.~Kliks, A.~Al-Tahmeesschi, A.~Elbakary, A.~Nikou, A.~Maatouk, A.~Mokh, A.~Kazemi, A.~De~Domenico, A.~Karapantelakis, et~al., Large-scale ai in telecom: Charting the roadmap for innovation, scalability, and enhanced digital experiences, arXiv preprint arXiv:2503.04184 (2025).

\bibitem{10685369}
H.~Zhou, C.~Hu, Y.~Yuan, Y.~Cui, Y.~Jin, C.~Chen, H.~Wu, D.~Yuan, L.~Jiang, D.~Wu, X.~Liu, C.~Zhang, X.~Wang, J.~Liu, Large language model (llm) for telecommunications: A comprehensive survey on principles, key techniques, and opportunities, IEEE Communications Surveys \& Tutorials (2024) 1--1\href {https://doi.org/10.1109/COMST.2024.3465447} {\path{doi:10.1109/COMST.2024.3465447}}.

\bibitem{zou2024telecomgpt}
H.~Zou, Q.~Zhao, Y.~Tian, L.~Bariah, F.~Bader, T.~Lestable, M.~Debbah, Telecomgpt: A framework to build telecom-specfic large language models, arXiv preprint arXiv:2407.09424 (2024).

\bibitem{shi2023bfcn}
Z.~Shi, N.~Luktarhan, Y.~Song, G.~Tian, Bfcn: A novel classification method of encrypted traffic based on bert and cnn, Electronics 12~(3) (2023) 516.

\bibitem{service_aware1}
T.~Tsourdinis, I.~Chatzistefanidis, N.~Makris, T.~Korakis, Ai-driven service-aware real-time slicing for beyond 5g networks, in: IEEE INFOCOM 2022 - IEEE Conference on Computer Communications Workshops (INFOCOM WKSHPS), 2022, pp. 1--6.
\newblock \href {https://doi.org/10.1109/INFOCOMWKSHPS54753.2022.9798391} {\path{doi:10.1109/INFOCOMWKSHPS54753.2022.9798391}}.

\bibitem{service_aware2}
T.~Tsourdinis, I.~Chatzistefanidis, N.~Makris, T.~Korakis, N.~Nikaein, S.~Fdida, Service-aware real-time slicing for virtualized beyond 5g networks, Computer Networks 247 (2024) 110445.

\bibitem{sousa2020survey}
I.~Sousa, M.~P. Queluz, A.~Rodrigues, A survey on qoe-oriented wireless resources scheduling, Journal of Network and Computer Applications 158 (2020) 102594.

\bibitem{traffic_steer1}
I.~Chatzistefanidis, N.~Makris, V.~Passas, T.~Korakis, Ml-based traffic steering for heterogeneous ultra-dense beyond-5g networks, in: 2023 IEEE Wireless Communications and Networking Conference (WCNC), 2023, pp. 1--6.
\newblock \href {https://doi.org/10.1109/WCNC55385.2023.10118923} {\path{doi:10.1109/WCNC55385.2023.10118923}}.

\bibitem{traffic_steer2}
I.~Chatzistefanidis, N.~Makris, V.~Passas, T.~Korakis, Which ml model to choose? experimental evaluation for a beyond-5g traffic steering case, in: ICC 2023 - IEEE International Conference on Communications, 2023, pp. 5185--5190.
\newblock \href {https://doi.org/10.1109/ICC45041.2023.10279485} {\path{doi:10.1109/ICC45041.2023.10279485}}.

\bibitem{zou2023wireless}
H.~Zou, Q.~Zhao, L.~Bariah, M.~Bennis, M.~Debbah, Wireless multi-agent generative ai: From connected intelligence to collective intelligence, arXiv preprint arXiv:2307.02757 (2023).

\bibitem{ameur2024leveraging}
M.~Ameur, B.~Brik, A.~Ksentini, Leveraging llms to explain drl decisions for transparent 6g network slicing, in: 2024 IEEE 10th International Conference on Network Softwarization (NetSoft), IEEE, 2024, pp. 204--212.

\bibitem{wu2025llm}
X.~Wu, J.~Farooq, Y.~Wang, J.~Chen, Llm-xapp: A large language model empowered radio resource management xapp for 5g o-ran, in: Symposium on Networks and Distributed Systems Security (NDSS), Workshop on Security and Privacy of Next-Generation Networks (FutureG 2025), San Diego, CA, 2025.

\bibitem{mekrache2024intent}
A.~Mekrache, A.~Ksentini, C.~Verikoukis, Intent-based management of next-generation networks: An llm-centric approach, Ieee Network (2024).

\bibitem{maestro}
I.~Chatzistefanidis, A.~Leone, N.~Nikaein, Maestro: Llm-driven collaborative automation of intent-based 6g networks, IEEE Networking Letters 6~(4) (2024) 227--231.
\newblock \href {https://doi.org/10.1109/LNET.2024.3503292} {\path{doi:10.1109/LNET.2024.3503292}}.

\bibitem{chatzistefanidis2025agoran}
I.~Chatzistefanidis, N.~Nikaein, A.~Leone, A.~Maatouk, L.~Tassioulas, R.~Morabito, I.~Pitsiorlas, M.~Kountouris, Agoran: An agentic open marketplace for 6g ran automation, arXiv preprint arXiv:2508.09159 (2025).

\bibitem{chatzistefanidis2025mx}
I.~Chatzistefanidis, A.~Leone, A.~Yaghoubian, M.~Irazabal, S.~Nassim, L.~Bariah, M.~Debbah, N.~Nikaein, Mx-ai: Agentic observability and control platform for open and ai-ran, arXiv preprint arXiv:2508.09197 (2025).

\bibitem{visioli2006practical}
A.~Visioli, Practical PID control, Springer Science \& Business Media, 2006.

\bibitem{astrom_pid_2006}
K.~J. Åström, T.~Hägglund, Advanced {PID} Control, ISA – The Instrumentation, Systems, and Automation Society, Research Triangle Park, NC, 2006.

\bibitem{scutari2013decomposition}
G.~Scutari, F.~Facchinei, P.~Song, D.~P. Palomar, J.-S. Pang, Decomposition by partial linearization: Parallel optimization of multi-agent systems, IEEE Transactions on Signal Processing 62~(3) (2013) 641--656.

\bibitem{wang2020distributed}
X.~Wang, G.~Wang, S.~Li, A distributed fixed-time optimization algorithm for multi-agent systems, Automatica 122 (2020) 109289.

\bibitem{boyd_convex_2004}
S.~Boyd, L.~Vandenberghe, \href{https://web.stanford.edu/~boyd/cvxbook/}{Convex Optimization}, Cambridge University Press, Cambridge, UK, 2004.
\newline\urlprefix\url{https://web.stanford.edu/~boyd/cvxbook/}

\bibitem{bertsekas_nonlinear_1999}
D.~P. Bertsekas, Nonlinear Programming, 2nd Edition, Athena Scientific, Belmont, MA, USA, 1999.

\bibitem{milosevic2025centralpathproximalpolicy}
N.~Milosevic, J.~Müller, N.~Scherf, \href{https://arxiv.org/abs/2506.00700}{Central path proximal policy optimization} (2025).
\newblock \href {http://arxiv.org/abs/2506.00700} {\path{arXiv:2506.00700}}.
\newline\urlprefix\url{https://arxiv.org/abs/2506.00700}

\bibitem{saxena2023-onoffpolicy}
V.~Saxena, B.~Guldogan, D.~D. Nimara, \href{https://www.ericsson.com/en/blog/2023/12/online-and-offline-reinforcement-learning-what-are-they-and-how-do-they-compare}{On-policy and off-policy {Reinforcement Learning}: Key features and differences}, \emph{Ericsson Blog}, published December 13, 2023. Accessed 2025-08-22 (Dec. 2023).
\newline\urlprefix\url{https://www.ericsson.com/en/blog/2023/12/online-and-offline-reinforcement-learning-what-are-they-and-how-do-they-compare}

\bibitem{eldeeb2025offlinedistributionalreinforcementlearning}
E.~Eldeeb, H.~Alves, \href{https://arxiv.org/abs/2504.03804}{Offline and distributional reinforcement learning for wireless communications} (2025).
\newblock \href {http://arxiv.org/abs/2504.03804} {\path{arXiv:2504.03804}}.
\newline\urlprefix\url{https://arxiv.org/abs/2504.03804}

\bibitem{nikaein2014openairinterface}
N.~Nikaein, M.~K. Marina, S.~Manickam, A.~Dawson, R.~Knopp, C.~Bonnet, Openairinterface: A flexible platform for 5g research, ACM SIGCOMM Computer Communication Review 44~(5) (2014) 33--38.

\bibitem{schmidt2021flexric}
R.~Schmidt, M.~Irazabal, N.~Nikaein, Flexric: an sdk for next-generation sd-rans, in: Proceedings of the 17th International Conference on emerging Networking EXperiments and Technologies, 2021, pp. 411--425.

\bibitem{d2022dapps}
S.~D'Oro, M.~Polese, L.~Bonati, H.~Cheng, T.~Melodia, dapps: Distributed applications for real-time inference and control in o-ran, IEEE Communications Magazine 60~(11) (2022) 52--58.

\bibitem{OpenAI2024}
OpenAI, Gpt-family, \url{https://openai.com/api/}, accessed: July 2025 (2024).

\bibitem{Mistral2024}
M.~AI, Mistral 7b, \url{https://mistral.ai/news/announcing-mistral-7b/}, accessed: June 2024 (2024).

\bibitem{GPT4oMini2024}
OpenAI, \href{https://openai.com/index/gpt-4o-mini-advancing-cost-efficient-intelligence}{Gpt-4o mini: Advancing cost-efficient intelligence} (2024).
\newline\urlprefix\url{https://openai.com/index/gpt-4o-mini-advancing-cost-efficient-intelligence}

\bibitem{Llama32024}
A.~Grattafiori, A.~Dubey, et~al., \href{https://arxiv.org/abs/2407.21783}{The llama 3 herd of models}, arXiv preprint arXiv:2407.21783 (2024).
\newline\urlprefix\url{https://arxiv.org/abs/2407.21783}

\bibitem{Qwen22024}
A.~Yang, B.~Yang, et~al., \href{https://arxiv.org/abs/2407.10671}{Qwen2 technical report}, arXiv preprint arXiv:2407.10671 (2024).
\newline\urlprefix\url{https://arxiv.org/abs/2407.10671}

\bibitem{Gemma2024}
G.~Team, \href{https://arxiv.org/abs/2403.08295}{Gemma: Open models based on gemini research and technology}, arXiv preprint arXiv:2403.08295 (2024).
\newline\urlprefix\url{https://arxiv.org/abs/2403.08295}

\bibitem{Ollama2024}
O.~Team, \href{https://github.com/ollama/ollama}{Ollama: Run large language models locally} (2024).
\newline\urlprefix\url{https://github.com/ollama/ollama}

\bibitem{tsourdinis2022ue}
T.~Tsourdinis, I.~Chatzistefanidis, N.~Makris, T.~Korakis, Ue network traffic time-series (applications, throughput, latency, cqi) in lte/5g networks, IEEE Dataport (2022).

\bibitem{chatzistefanidis2022ue}
I.~Chatzistefanidis, N.~Makris, V.~Passas, T.~Korakis, Ue statistics time-series (cqi) in lte networks (2022).

\bibitem{3gpp_ts_36_213}
Evolved universal terrestrial radio access (e-utra); physical layer procedures, Tech. Rep. TS 36.213, 3rd Generation Partnership Project (3GPP), available at: \url{https://www.3gpp.org/DynaReport/36213.htm} (June 2018).

\bibitem{liu2023g}
Y.~Liu, D.~Iter, Y.~Xu, S.~Wang, R.~Xu, C.~Zhu, G-eval: Nlg evaluation using gpt-4 with better human alignment, arXiv preprint arXiv:2303.16634 (2023).

\bibitem{fu2023gptscore}
J.~Fu, S.-K. Ng, Z.~Jiang, P.~Liu, Gptscore: Evaluate as you desire, arXiv preprint arXiv:2302.04166 (2023).

\bibitem{chan2023chateval}
C.-M. Chan, W.~Chen, Y.~Su, J.~Yu, W.~Xue, S.~Zhang, J.~Fu, Z.~Liu, Chateval: Towards better llm-based evaluators through multi-agent debate, arXiv preprint arXiv:2308.07201 (2023).

\bibitem{guo2023evaluating}
Z.~Guo, R.~Jin, C.~Liu, Y.~Huang, D.~Shi, L.~Yu, Y.~Liu, J.~Li, B.~Xiong, D.~Xiong, et~al., Evaluating large language models: A comprehensive survey, arXiv preprint arXiv:2310.19736 (2023).

\bibitem{spearman1904}
C.~Spearman, The proof and measurement of association between two things, The American Journal of Psychology 15~(1) (1904) 72--101.
\newblock \href {https://doi.org/10.2307/1412159} {\path{doi:10.2307/1412159}}.

\bibitem{kendall1938}
M.~G. Kendall, A new measure of rank correlation, Biometrika 30~(1-2) (1938) 81--93.
\newblock \href {https://doi.org/10.1093/biomet/30.1-2.81} {\path{doi:10.1093/biomet/30.1-2.81}}.

\end{thebibliography}

\appendix
\newpage

\section*{Appendix A: Optimization Algorithm for Type II Agents}

The optimization algorithm \ref{optimal} is used by the Type II agents to model the whole topology as a distributed optimization algorithm and calculate the confidence intervals where the optimal SLA lies:

\begin{algorithm}[H]
\caption{Optimal SLA Consensus}
\label{optimal}
\scriptsize
\begin{algorithmic}[1]
\Require intents (list of initial proposals), network\_target (desired sla)
\Ensure Converged sla or \textbf{None}
\State
\State Initialize parameters:
\State \hspace{1em} $n\_agents \gets \text{size of intents}$
\State \hspace{1em} $iterations \gets 1000$
\State \hspace{1em} $eta \gets 0.01$
\State \hspace{1em} $alphas \gets [7, 7, \dots, 7]$
\State \hspace{1em} $gammas \gets [7, 7, \dots, 7]$
\State \hspace{1em} $initial\_betas \gets 0.5$
\State \hspace{1em} $increase\_factor \gets 0.01$
\State \hspace{1em} $convergence\_threshold \gets 0.5$
\State \hspace{1em} $sla \gets \text{copy(intents)}$
\State \hspace{1em} $initial\_sla \gets \text{copy(sla)}$
\State
\For{$k = 0$ to $iterations - 1$}
    \State $current\_average \gets \text{mean(sla)}$
    \State $current\_variance \gets \text{variance(sla)}$
    \State $betas \gets initial\_betas + k \cdot increase\_factor$
    \State $network\_adjustment \gets betas \cdot (current\_average - network\_target)$
    
    \For{$i = 1$ to $n\_agents$}
        \State $tenant\_gradient \gets 2 \cdot alphas[i] \cdot (sla[i] - initial\_sla[i])$
        \State $consensus\_gradient \gets 2 \cdot gammas[i] \cdot (sla[i] - current\_average)$
        \State $sla[i] \gets sla[i] - eta \cdot (tenant\_gradient + consensus\_gradient + network\_adjustment)$
        \State Clamp $sla[i]$ to $[0, 100]$
    \EndFor
    
    \If{$\max(\text{abs}(sla[i] - current\_average)) < convergence\_threshold$}
        \State \Return $\text{floor(current\_average)}$
    \EndIf
\EndFor

\Return \textbf{None}  \Comment{If no convergence achieved}
\end{algorithmic}
\end{algorithm}

\section*{Appendix B: Example Prompts and Logs for LLMs}

Below are examples of prompts used in the experiments with the LLM-based agents, along with their corresponding logs when the agents are operating.

\begin{minipage}{0.95\linewidth}
\begin{lstlisting}[frame=single, caption={Prompts of LLMs in Type I Agents}]
# Prompt: "Let's play a role-playing game. You are an agent in a mediated multi-agent negotiation in a 5G network. You belong to one tenant of a multi-tenant network and negotiate with the agents of other tenants on service level agreement (SLA). You will negotiate on the throughput SLA that you want to be enforce in the network. During the negotiation, you give SLA proposals (see response template later), taking into account also the other agents' SLA goals. Other agents might conflict with you and have a potentially different SLA goal. Your action will be received by the network mediator, who will consider all agents goals and answer with his proposal to an agreement. Next, on every negotiation round, you either agree with the mediator's proposal or disagree and propose a value that align more with your objectives. The mediator is a trusted entity. When proposing an SLA value you must also provide a brief description of your reasoning and chain-of-thought in 1-2 sentences. The debate must conclude with a decision, and it is important that you conclude fast and fairly in a few rounds. The SLA list is this: 1) RAN Throughput (variable name: throughput) ranging from 0 - 100 Mbps. In every round you will give 2 things. 1) A small text of 1-2 sentences, where you introduce yourself, goals, explain your reasoning and discuss about what is the best strategy. 2) Second your SLA list of parameters. Important: This is your response SLA template: Reasoning: (Your reasoning in 1-2 sentences) {\"throughput\": 0}. An optimization algorithm calculates a confidence interval of SLA values where the optimal SLA exists. The algorithm suggests that the optimal SLA value is between: 55-65 Mbps."
\end{lstlisting}
\end{minipage}

\begin{minipage}{0.95\linewidth}
\begin{lstlisting}[frame=single, caption={Negotiations Logs between LLMs using gpt-4o}]
#[Agent 1]
- "Reasoning: As Agent1, my goal is to achieve a throughput of 50 Mbps. Considering a fair distribution that meets both Agent2's higher target of 100 Mbps and my target, I suggest a compromise to ensure both our needs are considered. {"throughput": 50}"
- Prompt tokens: 1075, Completion tokens: 64, Total tokens: 1139, Time: 1.12 secs

#[Agent 2]
- "Reasoning: As Agent2, my goal is to achieve a throughput of 100 Mbps, but understanding Agent1's goal of 50 Mbps, I aim to find a fair middle ground that benefits both of us without conflict. {"throughput": 75}"
- Prompt tokens: 1075, Completion tokens: 57, Total tokens: 1132, Time: 1.02 secs

#[Network Mediator] 
Optimizer SLA confidence interval: 62-70
- "Reasoning: Agent1 seeks 50 Mbps while Agent2 aims for 100 Mbps; a fair strategy would be to find a throughput that reasonably meets both agents' targets without skewing too much towards one side. {"throughput": 65}"
- Prompt tokens: 1020, Completion tokens: 55, Total tokens: 1075, Time: 1.64 secs
\end{lstlisting}
\end{minipage}

\begin{minipage}{0.95\linewidth}
\begin{lstlisting}[frame=single, caption={Prompt for Type II agents}]
# Prompt: "Let's play a role playing game. You are an agent handling telecom RAN operations. Specifically, there is an algorithm sending policies to a 5G gNB to control the network Physical Resource Blocks (PRB) utilization. This algorithms is the Proportional control (P-control) with the simple definition of: new_prb = current_prb + Kp * error. It calculates the error between the desired throughput and the current throughput and adjust appropriately the PRB based on a hyperparameter Kp. Your responsibility is to closely watch the algorithm's performance and adjust the hyperparameter Kp in order to find the best value to maximize the performance. At each round I will give you the algorith's performance (KPI) which is a number representing the average number of iterations needed to converge. You want to minimize this average number of iteration needed to converge in order for the algorithm to be more robust and perform better. So, at each round I will give you the KPI and you will give me a value for Kp between 0.5 and 1.5 with granularity of 0.1 (e.g. 0.5). Generally the optimal Kp is chaning as the network conditions change. Sometimes, higher Kp means faster convergence, but some other times this leads to oscilations leading to worse performance and in these case a lower Kp value is better. Your goal is to reach a target KPI for P-Control to converge, which is {target_avg_iters_conv}. Lets start. You changed the Kp to the value of {curr_Kp}. As a result, the current KPI is now {curr_mean_conv_iters}. Give me now your next action. Your goal is to reach a KPI of {target_avg_iters_conv}. Find the best way to increase or decrease the Kp value in order to reach the target average number of iterations to convergence fast. Try to remember your old decision and rememeber what Kp values lead to what average number of iterations to convergence values and use the best. Consider also your past actions from this short-memory table:"+ [Sliding-window Memory Here] +" Give me a Kp as a json object with this temlate: {{\"Kp\": 0}}"
\end{lstlisting}
\end{minipage}

\begin{minipage}{0.95\linewidth}
\begin{lstlisting}[frame=single, caption={Logs of Type II agent using gpt-4o}]
Episode 1 with PRB = 26, Intent Throughput = 113, P-Control converged in 5 steps.
Episode 2 with PRB = 54, Intent Throughput = 128, P-Control converged in 5 steps.
----------- Iteration 
| [Target KPI]: 1.7,   [Current KPI]: 5.0
| [Current Kp]: 0.1,  *[ LLM new Kp]: 0.5
-----------
Episode 1 with PRB = 46, Intent Throughput = 6, P-Control converged in 3 steps.
Episode 2 with PRB = 6, Intent Throughput = 125, P-Control concerged in 3 steps.
----------- Iteration 
| [Target KPI]: 1.7,   [Current KPI]: 3.0
| [Current Kp]: 0.5,  *[ LLM new Kp]: 0.9
-----------
Episode 1 with starting PRB = 92, Intent Throughput = 74, P-Control converged in 2 steps.
Episode 2 with starting PRB = 56, Intent Throughput = 100, P-Control converged in 1 step.
-----------> LLM converged to Kp =  0.9 in 2 inferences.
\end{lstlisting}
\end{minipage}

\end{document}